\documentclass{article}
\usepackage[accepted]{icml2026}
\usepackage{microtype}
\usepackage{graphicx}
\usepackage{booktabs}
\usepackage{multirow}
\usepackage{amsmath,amssymb}
\usepackage[version=4]{mhchem}
\usepackage{tcolorbox}
\tcbuselibrary{breakable}
\usepackage[capitalize,noabbrev]{cleveref}
\usepackage{tikz}
\usetikzlibrary{arrows.meta, positioning, calc}
\usetikzlibrary{positioning, fit, backgrounds}
\usepackage{threeparttable}
\usetikzlibrary{matrix}
\pgfdeclarelayer{background}
\pgfsetlayers{background,main}
\usepackage{enumitem}\usepackage{pifont}
\newcommand{\xmark}{\ding{55}}

\usepackage{float}

\newtcolorbox{tracebox}[2][]{
  breakable,
  colback=gray!4,
  colframe=black!55,
  boxrule=0.5pt,
  arc=2pt,
  left=6pt,
  right=6pt,
  top=6pt,
  bottom=6pt,
  title={#2},
  fonttitle=\bfseries,
  #1
}

\newtcolorbox{promptbox}[1][]{
  breakable,
  colback=blue!2,
  colframe=blue!35!black,
  boxrule=0.5pt,
  arc=2pt,
  left=6pt,
  right=6pt,
  top=6pt,
  bottom=6pt,
  #1
}

\definecolor{stage}{RGB}{220,230,242}
\definecolor{stageborder}{RGB}{100,130,170}
\definecolor{splitIID}{RGB}{218,238,218}
\definecolor{splitOODh}{RGB}{255,228,196}
\definecolor{splitOODp}{RGB}{255,218,218}
\definecolor{annot}{RGB}{120,120,120}
\definecolor{droptext}{RGB}{160,60,60}

\newcommand{\matpref}{\textsc{Mat-Pref}}

\icmltitlerunning{\matpref: Verifiable-Reward Training Improves Compositional Reasoning in Inorganic Materials}

\begin{document}
\twocolumn[
\icmltitle{\matpref: Verifiable-Reward Training Improves \\
           Compositional Reasoning in Inorganic Materials}

\begin{icmlauthorlist}
\icmlauthor{Sarrah Mikhail Leung}{berkeley,mlab}
\icmlauthor{Taehan Kim}{berkeley,mlab}
\icmlauthor{Jeongbin Park}{umich}
\end{icmlauthorlist}

\icmlaffiliation{berkeley}{University of California, Berkeley}
\icmlaffiliation{mlab}{Machine Learning at Berkeley}
\icmlaffiliation{umich}{University of Michigan}

\icmlcorrespondingauthor{Sarrah Mikhail Leung}{sarrahrose@berkeley.edu}

\icmlkeywords{Machine Learning, ICML, Materials Discovery, Reinforcement Learning, Verifiable Rewards}

\vskip 0.3in
]
\printAffiliationsAndNotice{}

\begin{abstract}
Reinforcement learning from verifiable rewards (RLVR) has driven rapid progress in mathematical and code reasoning, but when extended to science, existing benchmarks do not decompose what generalizes: do gains reflect structural transfer, property transfer, or memorization? We introduce \textsc{Mat-Pref}, a benchmark of 10,837 ionic-substitution questions across 11 inorganic structure families, grounded in density functional theory calculations from the Materials Project, with three evaluation splits that isolate in-distribution performance, generalization to entirely held-out structure families, and cross-property transfer: applying band-gap reasoning to hosts seen during training only through formation-energy supervision. Four zero-shot frontier models (70--671B parameters) remain in the 33--54\% range on every split, confirming that scale alone does not resolve the compositional chemical reasoning this task demands. A two-stage pipeline of supervised fine-tuning followed by Group Relative Policy Optimization (GRPO) lifts Qwen3-8B to 65.2\% in-distribution and 71.6\% on held-out families, exceeding zero-shot Qwen3-235B by over 20 percentage points on both structural-generalization splits, at a total pipeline cost under \$50. Self-consistency sampling shows that the SFT policy can already produce correct answers but cannot reliably surface them as the modal response; GRPO reshapes the distribution so that correct answers become modal rather than merely reachable, and this sharper commitment is visible mechanistically: logit lens analysis reveals a ${\sim}$20pp advantage in answer crystallization at the critical decision layer. We formalize this observation as a distractor-permutation consistency metric under which GRPO narrows the gap between lenient scoring (at least one permutation correct) and strict scoring (all permutations correct) from 24.0 to 14.3 percentage points. All results replicate on Llama-3.1-8B with identical hyperparameters, confirming the findings are not architecture-specific.

\end{abstract}

\section{Introduction}

Reinforcement learning from verifiable rewards (RLVR) has driven rapid progress in mathematical and code reasoning, and recent work has begun extending it to chemistry \citep{narayanan2025ether0}, biology \citep{istrate2025rbio1} and crystal structure generation \citep{hong2025generating}. In these scientific domains, correctness can often be verified through a computational database without the cost of synthesis or experiment. This positions simulation-derived properties as a form of world feedback: a reward signal grounded in physical law that scales beyond what laboratory measurement alone could provide. Existing benchmarks in this space do not decompose gains by generalization type, leaving open whether improvements reflect structural transfer, property transfer, or other mechanisms.  

\begin{figure*}[t]
\centering
\includegraphics[width=0.87\textwidth]{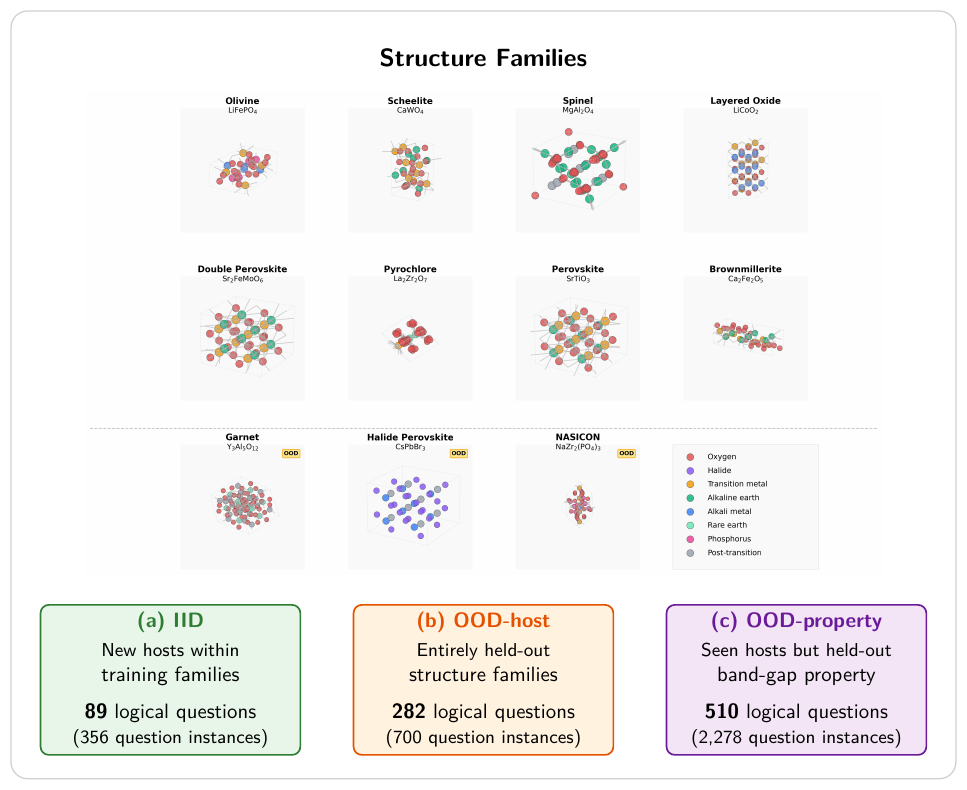}
\caption{%
Mat-Pref benchmark structure. Top: 11 inorganic structure families, with 8 used for training (above the dashed line) and 3 held out entirely for OOD-host evaluation (below). Bottom: three test splits. IID: novel hosts from training families. OOD-host: entirely held-out families (garnet, halide perovskite, NASICON). OOD-property: training-family hosts evaluated on the held-out band-gap property (template groups appeared in training only via formation-energy supervision). Counts are deduplicated logical questions (template group $\times$ property $\times$ direction $\times$ site) with raw question instances in parentheses.
}\label{fig:benchmark_overview}
\end{figure*}
Ionic substitution in inorganic crystals is a natural testbed: the correct substitution can be verified against Density Functional Theory (DFT) calculations, but identifying it requires jointly reasoning about oxidation states, coordination environments, ionic radii, and electronic-structure effects. We introduce \matpref{}, a benchmark of 10,837 multiple-choice ionic-substitution questions grounded in DFT properties from the Materials Project~\citep{jain2013materials}. Each question asks the model to select the best substitution from three or four candidates under a stated design goal. Every answer is backed by a first-principles calculation, enabling programmatic validation without synthesis or expert adjudication---a scalable reward signal for autonomous discovery.

\matpref{} probes generalization along three orthogonal axes
(Figure~\ref{fig:benchmark_overview}): to unseen hosts within
familiar structure families (IID), to entirely held-out structure families (OOD-host), and to a held-out target property for hosts seen during training only through a different property (OOD-property). Each underlying chemical question is evaluated across multiple distractor configurations, so accuracy reflects consistent reasoning rather than sensitivity to which alternatives accompany the correct answer. Current general-purpose models do not reliably compose the factors this benchmark requires: zero-shot accuracy across four frontier baselines spanning 70–235B parameters remains in the 33–54\% range on every split.

\paragraph{Prior work.}
Benchmarks for materials-science reasoning have evaluated supervised property prediction from crystal structure~\citep{dunn2020matbench}, natural-language comprehension of materials-science text~\citep{song2023matscinlp}, and LLM knowledge of crystal geometry~\citep{lv2025atomworld}. None test whether a language model can reason compositionally about substitution decisions whose correctness depends on joint structural, thermodynamic, and electronic considerations.
The closest methodological precedent is \citet{narayanan2025ether0}, who post-train a 24B model with
reinforcement learning on experimentally-grounded chemistry tasks verified via SMILES-based reward functions. Their ether0 model demonstrates that RLVR transfers to chemistry without domain pretraining, but targets organic small molecules where answers are molecular structures and evaluation reports held-out task accuracy without isolating what drives generalization across tasks. GRPO and related methods have been applied to mathematical reasoning~\citep{shao2024grpo}, code~\citep{guo2024deepseekcoder}, and inorganic crystal structure generation~\citep{hong2025generating}. These approaches target either open-ended generation or domains where verification is straightforward;  \matpref{} instead probes whether verifiable reward training can improve compositional scientific reasoning, where correctness depends on DFT-backed property rankings and evaluation splits isolate distinct axes of generalization. 

\paragraph{Contributions.}
This paper makes three contributions. First, we present \matpref{}, a DFT-grounded benchmark with three orthogonal evaluation axes, where each chemical question is scored by majority vote across its distractor permutations. Second, a two-stage pipeline of supervised fine-tuning followed by GRPO lifts Qwen3-8B ~\citep{yang2025qwen3} to 65.2\% IID accuracy and 71.6\% on entirely held-out structure families, exceeding the strongest zero-shot baseline we tested (Qwen3-235B-Instruct) by over 20 percentage points on both structural-generalization splits, with a smaller but positive margin on cross-property transfer. This result replicates on Llama-3.1-8B-Instruct with identical training data and hyperparameters (Table~\ref{tab:llama-replication}), confirming the finding 
generalizes across model families. 
Third, we show that GRPO produces more consistent answers than SFT across distractor permutations of the same underlying chemical question, narrowing the gap between lenient and strict scoring from  24.0pp to 14.3pp on average across all splits.

\Cref{sec:benchmark} describes the benchmark and evaluation protocol, \Cref{sec:training} summarizes the SFT+GRPO pipeline, \Cref{sec:experiments} presents the main results, and \Cref{sec:limitations} discusses scope and limitations.

\section{The \matpref{} Benchmark}
\label{sec:benchmark}
\begin{figure}[t]
\centering
\includegraphics[width=0.70\columnwidth]{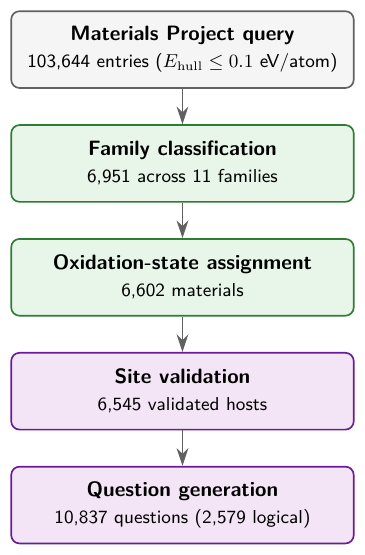}
\caption{\matpref{} construction pipeline. Entries are filtered by stability, classified into structure families (anonymous formula + anion + space-group filters), assigned oxidation states, validated at the site level (ChemEnv CN rules or ionic-radius fallback), and converted into multiple-choice questions with forward/reverse goals and gap-based filtering. The 10{,}837 questions are stratified by template group into IID, OOD-host, and OOD-property splits (Figure~\ref{fig:benchmark_overview}).}
\label{fig:pipeline}
\end{figure}

\matpref{} is built from the 103,644 Materials Project entries with energy above hull at most $0.1$~eV/atom, a standard threshold for experimentally synthesizable metastable phases~\citep{jain2013materials, sun2016metastability}. The construction pipeline (\Cref{fig:pipeline}) proceeds in four stages: family classification, oxidation-state assignment, site-level validation, and template-based question generation.

\paragraph{Family classification.}
Each entry is classified into one of 11 structure families using a three-stage procedure. The first stage matches anonymous formulas (e.g., \ce{ABC3}, \ce{AB2C4}), but these are inherently ambiguous: the same stoichiometric pattern can correspond to structurally distinct families. Two disambiguation stages resolve this: a compositional filter separates families by anion chemistry (oxide vs.\ halide), and space-group filters are applied to perovskite, halide perovskite, double perovskite, and NASICON, where stoichiometry and composition alone are insufficient (Section \ref{app:stats}, \Cref{tab:families}). After all three stages, 6,951 of the 103,644 entries are classified across the 11 target families.

\paragraph{Oxidation-state assignment and site validation.}
Formal oxidation states are assigned using the most probable charge-balance solution~\citep{ong2013pymatgen}, retaining 6,602 materials. Site assignments are derived from coordination environments computed by ChemEnv~\citep{waroquiers2020chemenv}. Cation coordination numbers are matched against family-specific rules: in perovskites, for example, cations with coordination number $\geq 8$ are assigned to the A-site and those with coordination number $\in \{5, 6, 7\}$ to the B-site. For layered oxides and double perovskites, where coordination environments do not distinguish sites, cations are sorted by Shannon ionic radius instead. Materials whose coordination numbers match no predefined rule are removed as likely misclassified, leaving 6,545 validated hosts. Full site-assignment rules are reported in \Cref{app:site_rules}.

\paragraph{Template-based question generation.}
Candidates are grouped by \emph{template composition}: the host formula with the substituted element removed. For example, \ce{CaTiO3} and \ce{BaTiO3} at the A-site share the template \texttt{?TiO$_3$}. Within a template group, all candidate materials share the same host framework and differ only at the substituted site, so property differences arise solely from the substituted element, rather than structural differences between hosts. 

For each template group with $N \geq 3$ distinct elements, questions are generated for both target properties (formation energy, band gap) and both goal directions. Forward questions ask the model to select the substitution that optimizes the property (most negative formation energy or largest band gap); reverse questions ask for the substitution that yields the worst value. The forward--reverse pairing tests whether the model reasons about chemical effects directionally rather than learning fixed element rankings. After deduplication, the benchmark contains 10,837 unique questions: 9,607 four-choice and 1,230 three-choice, split roughly evenly between forward (5,470) and reverse (5,367) goals and between formation energy (6,529) and band gap (4,308). 

The 10,837 questions correspond to 2,579 logical questions, each defined as a unique (template group, target property, goal direction, and crystallographic site) tuple. For example, ``which element should replace the A-site cation in \ce{NaCeO2} to maximize band gap?'' is one logical question. The same logical question may appear in the benchmark as multiple distinct four-choice questions that share the correct answer and runner-up but differ in which two additional distractors accompany them. The distribution is heavily skewed: 58.6\% of logical questions appear as a single four-choice question, while a long tail of frequently-substitutable template groups generates up to 105 permutations each (mean 4.2). Distractor permutations of a logical question are not statistically independent, so we report accuracy at the logical-question level throughout. A model that has learned the underlying chemistry answers all permutations consistently, while a model relying on surface cues will fluctuate. Question-level counts are provided in \Cref{tab:families} for comparability with standard multiple-choice evaluation.

\begin{figure*}[t]
\centering
\includegraphics[width=\textwidth]{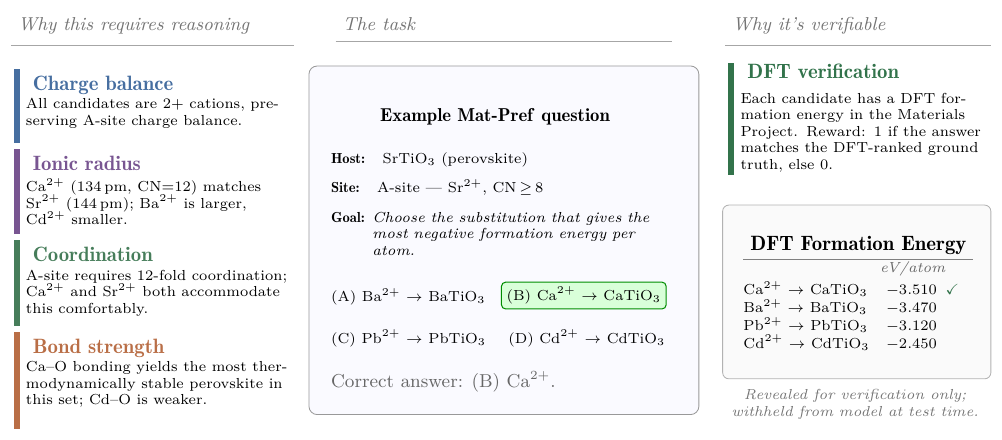}
\caption{\textsc{Mat-Pref} at a glance. \textbf{Middle:} each question specifies a host, a crystallographic site, a design goal, and three or four candidate substitutions; the model selects the best candidate and justifies its reasoning. \textbf{Left:} a correct choice composes several chemical principles, including charge balance, ionic radius, coordination preference, and bond strength, that no single heuristic captures. \textbf{Right:} every candidate has a first-principles formation energy in the Materials Project, so answers can be ranked and scored exactly without human preference labels. DFT values are shown here for illustration only and are withheld from the model.}
\label{fig:matpref_at_a_glance}
\end{figure*}

\subsection{Task design}

Each question specifies a host material, a crystallographic site, a
natural-language design goal, and three or four candidate
substitutions with their resulting compositions
(\Cref{fig:matpref_at_a_glance}). The model must select the candidate that best satisfies the stated goal and justify its reasoning.

The two target properties test distinct reasoning demands. Formation energy per atom measures thermodynamic stability, the primary screening objective in computational materials design, and is well-served by ionic-radius and charge-balance heuristics. Band gap probes electronic-structure reasoning: substitutions that alter $d$-orbital filling, covalency, or electronegativity contrast can shift a material from metallic to insulating, and no single heuristic reliably predicts the outcome. A model that performs well on both properties must compose thermodynamic and electronic arguments independently rather than rely on memorized element rankings.

Questions are retained only when the property gap between the best and second-best candidate exceeds a minimum threshold
$\delta_{\min}$ (0.02~eV/atom for formation energy, 0.10~eV for
band gap), ensuring an unambiguous correct answer. For four-choice questions, the answer set consists of the correct candidate, a competitive runner-up, and two distractors drawn from the remaining elements in the template group; three-choice questions include all elements. Answer positions are assigned by a deterministic shuffle to prevent position bias. Full selection criteria, difficulty classification, and edge-case statistics are reported in
\Cref{app:question_gen}.

\subsection{Evaluation splits}

The benchmark defines three evaluation axes, each isolating a
distinct form of generalization (\Cref{fig:benchmark_overview}).
For the structural splits (IID and OOD-host), all questions from a
given template group are assigned to the same split, preventing
group-level memorization. The OOD-property split operates on a
different axis: template groups from training families are tested
on the held-out band-gap property, so the same template group may
appear in both training (via formation energy) and OOD-property
testing (via band gap).

\paragraph{IID.}
Within the eight training families, template groups are split
90/5/5 into train, validation, and test. The IID test set contains 89 logical questions drawn from novel hosts within training families.

\paragraph{OOD-host.}
Three families are held out entirely (282 logical questions):
garnet, which introduces a complex oxide framework absent from
training; halide perovskite, which retains the perovskite topology but replaces oxide anions with halides; and NASICON, which introduces a phosphate-based polyanionic framework with distinct bonding motifs. These families were selected to isolate the effects of structural novelty, anion chemistry, and bonding character,
respectively.

\paragraph{OOD-property.}
This split contains 510 band-gap logical questions whose template
groups appear in training only through formation-energy
supervision. The model has never seen band-gap labels for any of
these hosts during training, so this axis tests whether chemical
reasoning learned from thermodynamic objectives transfers to
electronic-structure prediction on familiar frameworks.

\paragraph{Random baselines.}
Under majority voting at the logical-question level, the random baseline is 21.9\% (IID), 25.0\% (OOD-host), and 23.4\% (OOD-property). These baselines are more stringent than their question-level counterparts, since random guessing is less likely to win a majority across permutations than to answer any single permutation correctly. Detailed split statistics are reported in Section \ref{app:stats}, \Cref{tab:split-stats}.

\section{Training with Verifiable Rewards}
\label{sec:training}

Base models perform at or below chance on \matpref{} (\Cref{sec:experiments}), indicating that pretraining alone does not equip language models with the compositional reasoning required for ionic substitution. We adopt a two-stage pipeline: supervised fine-tuning to bootstrap the format and vocabulary of chemical reasoning, followed by GRPO to refine the policy against verifiable rewards.

\paragraph{SFT data generation.}
For each of the 7,287 training questions, a reasoning trace is generated by prompting DeepSeek-R1~\citep{deepseek2025r1} with the question and its correct answer, but without the underlying property values. The prompt instructs the model to reason from
chemical principles (ionic radius, coordination preference, electronegativity, bonding character) rather than numerical comparison. A three-step post-processing pipeline strips internal deliberation blocks, removes leaked numerical values via regular-expression matching, and verifies answer consistency with the ground truth. Of the 7,287 traces generated, 7,130 (97.8\%) pass all checks, with consistent yield across question types. Full prompts, post-processing details, and per-category yield are provided in \Cref{app:sft_traces}.

\paragraph{Supervised fine-tuning.}
Qwen3-8B~\citep{yang2025qwen3} is fine-tuned on the resulting (question, trace) pairs using LoRA~\citep{hu2022lora} (rank 32). The input follows the evaluation-time format: host material, site, design goal, and candidates are provided, but property values and the correct answer are withheld. Training runs for one epoch with a learning rate of $2 \times 10^{-4}$ and linear decay; the final evaluation loss (0.750) is below the training loss (0.797), consistent with no overfitting. SFT raises IID logical-question accuracy from 1.1\% to 44.9\%, a gain driven primarily by format learning, as the base model's extended-thinking mode exceeds the completion budget before producing a parseable answer.

\paragraph{GRPO.}
Starting from the SFT checkpoint, the model is trained with Group Relative Policy Optimization~\citep{shao2024grpo} on the same 7,287 questions. For each batch of 64 questions, 8 completions are sampled at temperature 0.8 and scored with a binary reward: 1 if the extracted answer matches the DFT-backed ground truth, 0 otherwise. Rewards are normalized within each group, and constant-reward groups are discarded. This exact-match reward is feasible because every candidate has a DFT-computed property value, making the ranking unambiguous.

\begin{figure}[t]
  \centering
  \includegraphics[width=1.0\columnwidth]{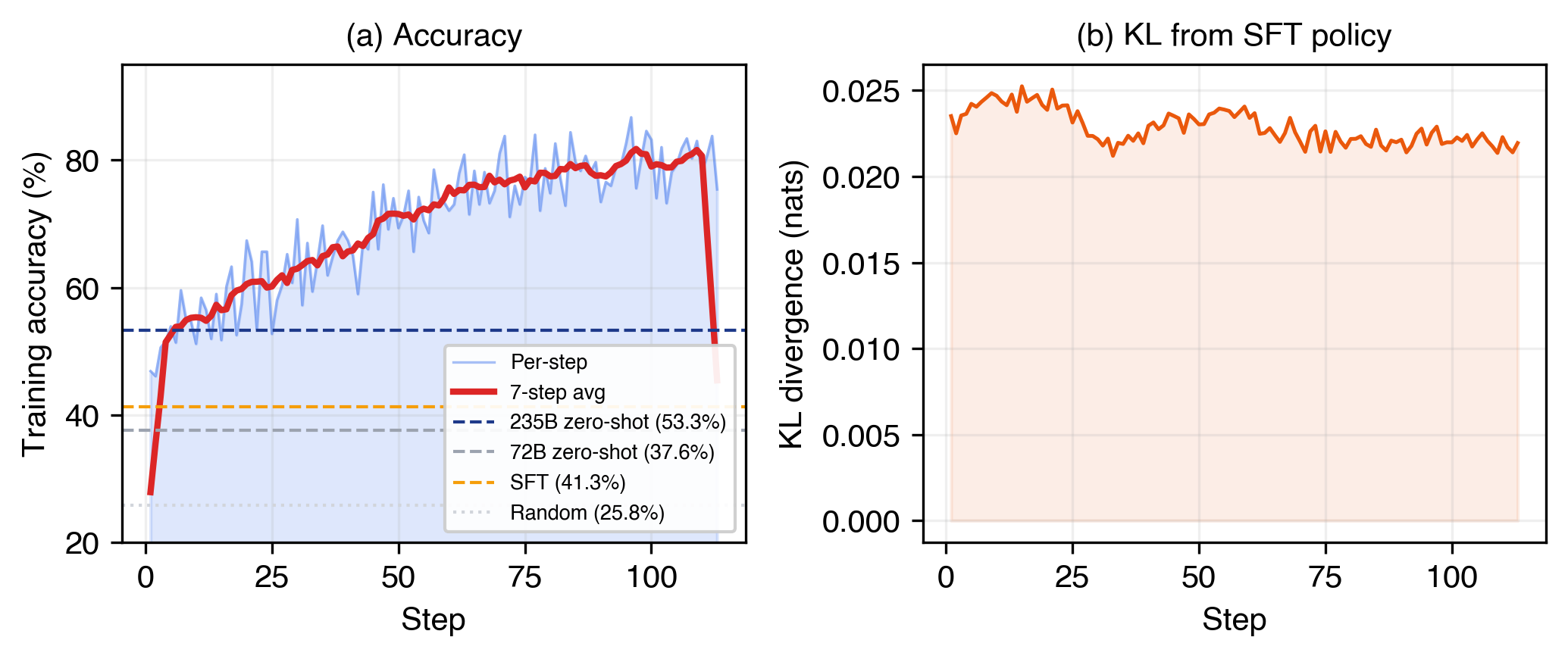}
  \caption{GRPO training dynamics.
  (a)~Question-level training accuracy rises from 55\% to 80\%
  (7-step moving average). Dashed lines show reference baselines:
  random (25.8\%), Qwen2.5-72B zero-shot (37.6\%), SFT (41.3\%),
  and Qwen3-235B zero-shot (53.3\%).
  (b)~KL divergence from the SFT reference policy stays below
  0.025 nats. Response length and entropy stability are reported in
  \Cref{fig:dynamics_full}.}
  \label{fig:dynamics}
\end{figure}

Training runs for 113 steps (one epoch). Training accuracy rises from 55.0\% to 80.3\% (\Cref{fig:dynamics}a), and KL divergence from the SFT reference stays below 0.025 nats throughout (\Cref{fig:dynamics}b), confirming that GRPO refines the SFT
policy without catastrophic drift. Full training dynamics and hyperparameters are reported in \Cref{app:hyper}. Response length and policy entropy remain stable throughout (\Cref{fig:dynamics_full}).

\section{Experiments}
\label{sec:experiments}

\begin{table}[t]
\centering
\caption{Logical-question accuracy (\%) across training stages and evaluation splits. Random baselines are computed under majority voting across distractor permutations. Wilson 95\% CIs for GRPO: IID [54.8, 74.3], OOD-host [66.1, 76.6], OOD-property [55.7, 64.2].}

\label{tab:results}
\scriptsize
\setlength{\tabcolsep}{2.5pt}
\begin{tabular*}{\columnwidth}{@{\extracolsep{\fill}}lccc@{}}
\toprule
Model & IID & OOD-host & OOD-prop. \\
\midrule
Random baseline & 21.9\% & 25.0\% & 23.4\% \\
Llama 3.3-70B & 33.7\% & 35.1\% & 46.3\% \\
DeepSeek-V3 & 36.0\% & 36.5\% & 50.6\% \\
Qwen2.5-72B & 37.1\% & 33.3\% & 39.6\% \\
Qwen3-235B-Instruct & 43.8\% & 45.4\% & 53.3\% \\
\midrule
Qwen3-8B + SFT & 44.9\% & 47.9\% & 50.2\% \\
Qwen3-8B + SFT (SC@8) & 50.6\% & 50.4\% & 51.8\% \\
Qwen3-8B + SFT + GRPO & \textbf{65.2\%} & \textbf{71.6\%} & \textbf{60.0\%} \\
\bottomrule
\end{tabular*}

\end{table}

\subsection{Baselines}
\label{sec:baselines}
We evaluate four zero-shot frontier baselines spanning 70--671B parameters to
characterize benchmark difficulty (\Cref{tab:results}): Qwen2.5-72B~\citep{yang2025qwen25},
Llama~3.3-70B~\citep{grattafiori2024llama3}, DeepSeek-V3~\citep{deepseekai2024deepseekv3},
and Qwen3-235B-Instruct~\citep{yang2025qwen3}. Qwen3-235B is the strongest zero-shot model on every split, but remains in the 43--53\% range (\Cref{tab:results}), confirming that model scale does not resolve the compositional reasoning \matpref{} requires. The controlled comparison is SFT vs.\ SFT+GRPO on the same 8B architecture, which isolates the effect of verifiable-reward training from the contribution of the underlying model. Base Qwen3-8B's 3.9\% IID accuracy (question-level) reflects completion-budget truncation (82.6\% parse failures) rather than reasoning inability; SFT resolves this entirely, showing that the two-stage pipeline addresses format and reasoning as separable bottlenecks.

\subsection{Two-stage training lifts accuracy above frontier baselines}
\label{sec:training_results}

SFT raises IID logical-question accuracy from 1.1\% to 44.9\%, with comparable gains on OOD-host (47.9\%) and OOD-property (50.2\%).  GRPO lifts IID accuracy to 65.2\% and OOD-host to 71.6\%, exceeding the strongest zero-shot baseline (Qwen3-235B-Instruct) by over 20 percentage points on both structural-generalization splits and by 6.7pp on OOD-property. Across all these splits, GRPO fixes 213 logical questions that SFT answered incorrectly while introducing 78 regressions, a fix-to-regress ratio of 2.73$\times$ (full breakdown in \Cref{app:question_flow}). 

To test whether GRPO's gains reflect policy improvement rather than better search over a flat SFT distribution, we evaluate self-consistency with 8 samples (SC@8) on the SFT checkpoint at $T{=}0.8$, matching GRPO's training-time sampling.  SC@8 closes only 11--28\% of the GRPO--SFT gap across splits (Table~\ref{tab:results}). The SFT policy can produce the correct answer in at least one of eight samples roughly 90\% of the time, but cannot surface it as the modal response; GRPO reshapes the policy to make correct answers modal rather than merely reachable.

\subsection{Results replicate across model families}
\label{sec:replication}

To rule out the possibility that GRPO's gains are an artifact of 
Qwen3-8B specifically, we replicate the SFT + GRPO pipeline on 
Llama-3.1-8B-Instruct using identical training data, 
hyperparameters, and evaluation protocol. The pattern reproduces 
cleanly at the logical-question level 
(Table~\ref{tab:llama-replication}): GRPO improves over SFT by 
$+21.4$pp on IID, $+15.6$pp on OOD-host, and $+4.7$pp on 
OOD-property, matching the qualitative structure of the Qwen3-8B 
result on every split. Notably, GRPO accuracy on OOD-property 
converges to 60.0\% for both model families, suggesting that 
cross-property transfer is structurally bounded at this dataset 
size rather than limited by model-specific representations.

\begin{table}[t]
\centering
\caption{Replication on Llama-3.1-8B-Instruct with identical 
training data and hyperparameters. Logical-question accuracy 
(\%, majority vote) with Wilson 95\% CIs.}
\label{tab:llama-replication}
\small
\begin{tabular}{@{}lccccc@{}}
\toprule
Split & $N$ & Base & SFT & GRPO & $\Delta$ \\
\midrule
IID          &  89 & 18.0 & 49.4 & \textbf{70.8} & +21.4 \\
OOD-host     & 282 & 27.3 & 57.8 & \textbf{73.4} & +15.6 \\
OOD-property & 510 & 27.8 & 55.3 & \textbf{60.0} & +4.7 \\
\bottomrule
\end{tabular}
\end{table}

\subsection{GRPO stabilizes reasoning across distractor permutations}
\label{sec:consistency}

In addition to answering more questions correctly, GRPO answers them more consistently across distractor permutations. We quantify this as the gap between \emph{lenient} scoring (the model answers at least one distractor permutation correctly) and \emph{strict} scoring (the model answers all permutations correctly). SFT shows a 24.0pp average lenient--strict gap across splits; GRPO shrinks this to 14.3pp. The reduction is significant under paired cluster bootstrap on all three splits: IID +14.6pp [+3.3, +26.4], OOD-host +7.1pp [+3.2, +11.4], OOD-property +7.5pp [+4.4, +10.6].

This effect is only visible when permutations of the same underlying question are grouped; question-level scoring treats each permutation independently and averages over the inconsistency. The mean fraction of permutations answered correctly per logical question tells the same story: GRPO exceeds SFT on all splits (full results in \Cref{app:consistency}).

Logit lens analysis~\citep{belrose2023tuned} offers a mechanistic explanation: projecting intermediate residual streams through the unembedding matrix reveals that both models undergo a sharp answer-decision transition at layer 24, but GRPO's crystallization rate at that layer is 79.6\% vs.\  SFT's 60.2\%, with the $\sim$20pp advantage sustained through the final layer. This suggests that GRPO's consistency gains  stem from sharper internal commitment to the selected answer rather than earlier decision-making (\Cref{app:logit_lens}).

\subsection{Electronic structure reasoning as the primary bottleneck}
\label{sec:per_property}

\begin{table}[t]
\centering
\caption{GRPO logical-question accuracy (\%) by target property
and goal direction across evaluation splits.}
\label{tab:per_property}
\footnotesize
\setlength{\tabcolsep}{2pt}
\begin{tabular}{@{}lcccc@{}}
\toprule
 & \multicolumn{2}{c}{Property}
 & \multicolumn{2}{c}{Direction} \\
\cmidrule(lr){2-3} \cmidrule(lr){4-5}
Split & Form.\ en. & Band gap & Forward & Reverse \\
\midrule
IID          & 71.2 & 47.8 & 61.4 & 68.9 \\
OOD-host     & 69.9 & 73.6 & 74.3 & 69.2 \\
OOD-property & ---  & 60.0 & 60.0 & 60.0 \\
\bottomrule
\end{tabular}
\end{table}

\Cref{tab:per_property} reveals two asymmetries. Formation-energy accuracy remains above 69\% on every split while band-gap accuracy ranges from 47.8\% (IID) to 73.6\% (OOD-host); we analyse this in \Cref{sec:ood_generalization}. Two confounds apply: the training set contains $\sim$3.6$\times$ more formation-energy than band-gap questions (\Cref{tab:split-stats}), and PBE-DFT band gaps carry larger systematic error than formation energies. 

Second, the forward--reverse asymmetry is not consistent across splits. On IID, reverse-goal questions are answered 7.5pp more accurately than forward-goal questions, but on OOD-host the direction reverses (forward +5.1pp over reverse), and on OOD-property the two are balanced (both 60.0\%). This inconsistency contrasts with the generally reverse $>$ forward pattern seen across all four zero-shot baselines (\Cref{tab:zero-shot-by-direction}). Zero-shot models likely benefit on reverse-goal questions because the worst candidate is often an obvious outlier; GRPO's lack of consistent direction bias suggests it reasons about the design goal rather than defaulting to this heuristic. 

\subsection{Generalization to unseen structure families}
\label{sec:ood_generalization}

\begin{table}[t]
\centering
\caption{Per-family logical-question accuracy on the OOD-host split. Deltas use paired cluster bootstrap (B=10,000), resampling template groups. Overall delta +26.2pp [+19.2, +33.1].}
\label{tab:ood_host_family}

\begin{tabular*}{\columnwidth}{@{\extracolsep{\fill}}lcccc@{}}
\toprule
Family & $N$ & GRPO & Qwen3-235B & $\Delta$ \\
\midrule
NASICON        &  28 & 82.1\% & 50.0\% & +32.1pp \\
Garnet         & 171 & 71.3\% & 43.3\% & +28.0pp \\
Halide perov.  &  83 & 68.7\% & 48.2\% & +20.5pp \\
\midrule
\textbf{Total} & 282 & \textbf{71.6\%} & 45.4\% & \textbf{+26.2pp} \\
\bottomrule
\end{tabular*}
\end{table}

GRPO outperforms zero-shot Qwen3-235B-Instruct on every held-out family (\Cref{tab:ood_host_family}). NASICON achieves the highest OOD accuracy (82.1\%), suggesting that ionic-substitution heuristics transfer even to polyanionic frameworks absent from training.  Halide perovskite scores lowest (68.7\%) despite sharing the perovskite topology with training data, indicating that changes in anion chemistry pose greater difficulty than changes in framework geometry. Garnet (71.3\%) performs consistently across both properties. Within halide perovskite, per-anion 
accuracy (\Cref{tab:halide-by-anion}) is heterogeneous: sample sizes are too small to support directional claims about chemical similarity to the oxide training distribution.

OOD-host accuracy (71.6\%) exceeds IID (65.2\%) but this inversion is driven entirely by band-gap (\Cref{tab:per_property}). Formation-energy accuracy is 
stable across splits (71.2\% IID vs.\ 69.9\% OOD-host), while band-gap accuracy jumps from 47.8\% on IID to 73.6\% on OOD-host. The two splits differ in element composition (\Cref{tab:bg-element-categories}): IID band-gap questions are dominated by transition-metal substitutions (52\%) where d-orbital effects govern the answer, while OOD-host band-gap questions shift the probability mass toward post-transition metals such as Al, Ga and In (21\% of OOD-host LQs, absent from IID), which lack d-electrons and whose band-gap effects are predictable from size and electronegativity alone. The inversion therefore reflects the model's relative strength on size and electronegativity reasoning over d-orbital effects, rather than superior generalization to unseen families.

\subsection{Formation-energy training transfers to band-gap reasoning}
\label{sec:cross_property}

OOD-property asks whether formation-energy training teaches the model anything about band gaps. GRPO reaches 60.0\% on 510 logical questions, seen only through formation-energy supervision, above both the random baseline (23.4\%) and every zero-shot baseline we tested, including Qwen3-235B-Instruct (53.3\%). The margin over the strongest baseline is the smallest across the three splits, suggesting that frontier pretraining captures cross-property reasoning but does not close the gap. Per-family accuracy is heterogeneous: perovskite reaches 72.0\% while brownmillerite falls to 35.7\%, consistent with families where ionic-radius heuristics predict band-gap trends transferring better than those where electronic structure effects dominate (\Cref{tab:ood_property_family}).


\section{Scope and Limitations}
\label{sec:limitations}
\paragraph{Benchmark and task scope.}
\matpref{} is multiple choice over a finite candidate set defined by Materials Project coverage. It does not test open-ended materials generation, synthesis feasibility, or laboratory decision-making. All ground-truth answers are derived from DFT values computed at 0\,K, so the benchmark inherits the approximations of that data source, including the PBE exchange-correlation functional, the neglect of finite-temperature entropic contributions, and incomplete coverage of the chemical space. Performance on \matpref{} should therefore be understood as reasoning ability within a DFT-consistent framework, not as predictive accuracy for real synthesis outcomes. The template-based design ensures scientific validity but produces a smaller benchmark than approaches that pool candidates across unrelated materials, and the relaxed gap thresholds ($\delta_{\min}$ = 0.02 ~eV/atom for formation energy, 0.10~eV for band gap) admit questions where the correct answer is separated from the runner-up by margins smaller than typical DFT uncertainty, though 87.5\% of questions exceed twice the minimum threshold. 

\paragraph{Effective sample size.}
The test set contains 881 logical questions (89 IID, 282 OOD-host, and 510 OOD-property). IID is the smallest split, and its band-gap subset (23 logical questions) is too small for conclusive cross-property comparisons.  Question-level numbers are provided in \Cref{app:question_level}. 

\paragraph{Training and future directions.}
Training and evaluation use two 8B-class instruction-tuned models (Qwen3-8B and Llama-3.1-8B-Instruct); extending to larger base models or non-instruction-tuned checkpoints remains future work. 
Only two target properties are included; extending to elastic moduli or dielectric constants would test whether verifiable-reward training generalizes beyond thermodynamic and electronic reasoning. 
The analysis in \Cref{sec:per_property} identifies electronic-structure reasoning as the primary bottleneck: IID band-gap accuracy lags formation energy by over 20 percentage points. This gap likely reflects a limitation of electronic structure reasoning from a binary reward alone, without supervision on the intermediate chemical arguments that connect substitution to band-gap change. Process reward models that score intermediate reasoning steps, or compositional rewards that separately assess ionic-radius matching and electronic-structure arguments, represent natural directions for future work. 


\section{Evaluation beyond \matpref{}}
\label{sec:transfer}
To probe whether training on \matpref{} improves reasoning beyond its curated format, we evaluate both base Qwen3-8B and the GRPO-trained checkpoint on the 42 text-only chemistry questions (15 multiple-choice, 27 exact-match) from the HLE-Gold Bio/Chem subset~\citep{futurehouse2025hlegold}, a validated collection we use after recent analyses questioned a substantial fraction of the original HLE chemistry/biology items~\citep{futurehouse2025hlewrong}. Both models are queried under identical decoding settings with a 4,096-token completion budget. On multiple-choice questions, the trained model reaches 5/15 (33.3\%) versus 2/15 (13.3\%) for the base; on exact-match, 1/27 versus 0/27. The sample is too small for strong accuracy claims, but a consistent qualitative difference emerges: the trained model produces concise, decisive responses, while the base model often generates long, meandering traces that fail to terminate cleanly (see \Cref{app:hle-gold-traces}). This pattern suggests that both chemical reasoning and the structural response behavior learned through \matpref{} transfer to some degree, though the sample is too small to disentangle their contributions.

\section{Conclusion}

\matpref{} provides a controlled testbed for verifiable-reward training in materials science, with DFT-backed answers and evaluation splits that isolate structural generalization and cross-property transfer. SFT followed by GRPO lifts Qwen3-8B above all zero-shot models we tested --- including models 30$\times$ larger --- on every split, while producing more consistent reasoning across distractor permutations of the same underlying chemistry. 

These findings point to concrete next steps. First, electronic-structure reasoning lags thermodynamic reasoning by over 20 percentage points on IID, where transition-metal substitutions dominate, motivating process reward models that score intermediate reasoning steps rather than final answers alone. Second, the compressed margin on OOD-property suggests that cross-property transfer is the regime where frontier pretraining most closely approaches task-specific training, and that richer reward structures than binary exact-match may be needed to push further. Third, the consistency gain motivates rewards that explicitly penalize instability across surface-form perturbations. Scaling the logical-question test pool, extending to additional target properties, and validating against synthesis databases would test whether the reasoning learned here transfers beyond the DFT-consistent setting in which it was trained.

\section*{Code \& Data Availability}
Code and datasets are provided at https://github.com/sarrahroseml/mat-pref. 

\bibliographystyle{plainnat}
\bibliography{references}

\clearpage
\onecolumn
\appendix

\section{Additional Dataset Statistics}
\label{app:stats}

\begin{table}[H]
\centering
\begin{threeparttable}
\caption{Evaluation split statistics}
\label{tab:split-stats}
\begin{tabular}{lcccccc}
\toprule
Split & Questions & TGs & Qs/group & FE & BG & 3-choice \\
\midrule
train & 7,287 & 581 & 12.5 & 5,695 & 1,592 & 742 \\
val & 216 & 32 & 6.8 & 184 & 32 & 43 \\
test iid & 356 & 33 & 10.8 & 273 & 83 & 33 \\
test ood host & 700 & 79 & 8.9 & 377 & 323 & 153 \\
test ood property & 2,278 & 276 & 8.3 & 0 & 2,278 & 259 \\
\midrule
Total & 10,837 & 723\tnote{a} & 15.0 & 6,529 & 4,308 & 1,230 \\
\bottomrule
\end{tabular}
\begin{tablenotes}\footnotesize
\item[a] Per-split TG counts sum to 725 because two stoichiometric template-group labels (\texttt{?-Li12-O48-P12} and \texttt{?-Li6-O24-P6}) appear in both an olivine host (training splits) and a garnet host (OOD-host split). The total reports the deduplicated union. Held-out isolation is enforced at the (template group, structure family) pair level.
\end{tablenotes}
\end{threeparttable}
\end{table}

\begin{table}[H]
\centering
\caption{Element-category distribution of correct answers in band-gap logical questions, by evaluation split. Post-transition metals (Al, Ga, In, Tl, Sn, Pb, Bi) are absent from IID band-gap but account for 20.9\% of OOD-host band-gap LQs.}
\label{tab:bg-element-categories}
\small
\begin{tabular}{@{}lcc@{}}
\toprule
Category & IID band gap & OOD-host band gap \\
 & ($n=23$ LQs) & ($n=129$ LQs) \\
\midrule
Transition metal     & 52.2\% (12) & 47.3\% (61) \\
Rare earth           & 26.1\% (6)  & 14.0\% (18) \\
Post-transition metal & 0.0\% (0)  & 20.9\% (27) \\
Alkaline earth       & 8.7\% (2)   & 11.6\% (15) \\
Alkali metal         & 8.7\% (2)   & 5.4\% (7)   \\
Other                & 4.3\% (1)   & 0.8\% (1)   \\
\bottomrule
\end{tabular}
\end{table}

\begin{table}[H]
\centering
\small
\begin{threeparttable}
\caption{Structure families in \matpref{}. \textbf{Filters:} \textsc{o} = requires oxygen; \textsc{halide} = halide anions; \textsc{sg} = space-group filter. \textbf{Site method:} CN = coordination number from ChemEnv; $\dagger$ = ionic-radius sorting fallback. OOD families are held out for out-of-distribution evaluation.}
\label{tab:families}
\begin{tabular}{lllcrrrr}
\toprule
Family & Sites & Filters & Site & Materials & TGs & Qs/grp & Questions \\
\midrule
Layered oxide$^\dagger$     & A, B       & \textsc{o}              & radius &    729 & 102 & 21.0 & 2{,}146 \\
Perovskite                  & A, B       & \textsc{o}, \textsc{sg} & CN     &    511 & 102 & 18.9 & 1{,}927 \\
Scheelite                   & A, B       & \textsc{o}              & CN     &    882 &  96 & 19.3 & 1{,}856 \\
Spinel                      & A, B       & \textsc{o}              & CN     &  1{,}157 & 130 & 13.5 & 1{,}757 \\
Pyrochlore                  & A, B       & \textsc{o}              & CN     &    515 &  86 & 12.4 & 1{,}068 \\
Double perov.$^\dagger$     & A, B, B$'$ & \textsc{o}, \textsc{sg} & radius &    615 &  84 & 12.2 & 1{,}029 \\
Olivine                     & M, T       & \textsc{o}              & CN     &  1{,}236 &  43 &  5.7 &    244 \\
Brownmillerite              & A, B       & \textsc{o}              & CN     &    206 &  21 &  5.2 &    110 \\
\midrule
Garnet\textsubscript{\scriptsize OOD}        & A, B, C    & \textsc{o}                   & CN &    444 &  48 & 10.2 &    491 \\
Halide perov.\textsubscript{\scriptsize OOD} & A, B       & \textsc{halide}, \textsc{sg} & CN &    180 &  23 &  7.2 &    165 \\
NASICON\textsubscript{\scriptsize OOD}       & A, B, C    & \textsc{o}, \textsc{sg}      & CN &     70 &   8 &  5.5 &     44 \\
\midrule
Total                       & & & &  6{,}545 & 723\tnote{a} & 15.0 & 10{,}837 \\
Training families (8)       & & & &  5{,}851 & 646\tnote{a} & --- & 10{,}137 \\
OOD-host families (3)       & & & &    694 &  79 & --- &    700 \\
\bottomrule
\end{tabular}
\begin{tablenotes}\footnotesize
\item[a] Per-family TG counts sum to 664 (training) and 727 (all 11) because 18 stoichiometric template-group labels appear in more than one structure family (e.g.\ \texttt{?-Mn4-O8} crystallizes as both a layered manganate and a spinel). Totals report the deduplicated union. Held-out isolation is enforced at the (template group, structure family) pair level.
\end{tablenotes}
\end{threeparttable}
\end{table}

\begin{table}[H]
\centering
\caption{Per-family logical-question accuracy (\%) on the
OOD-property split. All questions target band gap on template
groups seen in training only through formation-energy supervision.}
\label{tab:ood_property_family}
\begin{tabular}{lcc}
\toprule
Family & $N_{\text{logical}}$ & GRPO accuracy \\
\midrule
Spinel & 129 & 63.6\% [55.0, 71.4] \\
Scheelite & 85 & 56.5\% [45.9, 66.5] \\
Perovskite & 75 & 72.0\% [61.0, 80.9] \\
Pyrochlore & 69 & 65.2\% [53.4, 75.4] \\
Layered oxide & 56 & 46.4\% [34.0, 59.3] \\
Olivine & 41 & 68.3\% [53.0, 80.4] \\
Double perovskite & 41 & 43.9\% [29.9, 59.0] \\
Brownmillerite & 14 & 35.7\% [16.3, 61.2] \\
\midrule
Total & 510 & 60.0\% [55.7, 64.2] \\
\bottomrule
\end{tabular}
\end{table}

\subsection{Zero-Shot Baseline Details}
\label{app:baseline_details}

Tables~\ref{tab:zero-shot-by-property}--\ref{tab:zero-shot-by-family} 
report per-property, per-direction, and per-family breakdowns for 
all three zero-shot baselines at the logical-question level.

\begin{table}[H]
\centering
\caption{Zero-shot logical-question accuracy by target property, with Wilson 95\% CIs.}
\label{tab:zero-shot-by-property}
\begin{tabular}{llcc}
\toprule
Model & Split & Formation energy & Band gap \\
\midrule
Qwen2.5-72B & iid & 36.4\% [25.8, 48.4] (n=66) & 39.1\% [22.2, 59.2] (n=23) \\
Qwen2.5-72B & ood-host & 30.1\% [23.4, 37.7] (n=153) & 37.2\% [29.4, 45.8] (n=129) \\
Qwen2.5-72B & ood-property & --- & 39.6\% [35.5, 43.9] (n=510) \\
\midrule
Llama 3.3-70B & iid & 27.3\% [18.0, 39.0] (n=66) & 52.2\% [33.0, 70.8] (n=23) \\
Llama 3.3-70B & ood-host & 30.7\% [24.0, 38.4] (n=153) & 40.3\% [32.2, 48.9] (n=129) \\
Llama 3.3-70B & ood-property & --- & 46.3\% [42.0, 50.6] (n=510) \\
\midrule
DeepSeek-V3 & iid & 31.8\% [21.8, 43.8] (n=66) & 47.8\% [29.2, 67.0] (n=23) \\
DeepSeek-V3 & ood-host & 34.6\% [27.6, 42.5] (n=153) & 38.8\% [30.8, 47.4] (n=129) \\
DeepSeek-V3 & ood-property & --- & 50.6\% [46.3, 54.9] (n=510) \\
\midrule
Qwen3-235B & iid & 39.4\% [28.5, 51.5] (n=66) & 56.5\% [36.8, 74.4] (n=23) \\
Qwen3-235B & ood-host & 37.3\% [30.0, 45.1] (n=153) & 55.0\% [46.4, 63.4] (n=129) \\
Qwen3-235B & ood-property & --- & 53.3\% [49.0, 57.6] (n=510) \\
\bottomrule
\end{tabular}
\end{table}

\begin{table}[H]
\centering
\caption{Zero-shot logical-question accuracy decomposed by direction (forward = maximize property, reverse = minimize). Wilson 95\% CIs.}
\label{tab:zero-shot-by-direction}
\small
\begin{tabular}{llcc}
\toprule
Model & Split & Forward & Reverse \\
\midrule
Qwen2.5-72B & iid & 27.3\% [16.3, 41.8] (12/44) & 46.7\% [32.9, 60.9] (21/45) \\
Qwen2.5-72B & ood-host & 26.5\% [19.8, 34.4] (36/136) & 39.7\% [32.2, 47.8] (58/146) \\
Qwen2.5-72B & ood-property & 36.4\% [31.0, 42.2] (102/280) & 44.9\% [37.2, 49.9] (100/230) \\
\midrule
Llama 3.3-70B & iid & 31.8\% [19.9, 46.6] (14/44) & 35.6\% [23.2, 50.2] (16/45) \\
Llama 3.3-70B & ood-host & 25.7\% [19.1, 33.6] (35/136) & 43.8\% [36.1, 51.9] (64/146) \\
Llama 3.3-70B & ood-property & 40.4\% [34.8, 46.2] (113/280) & 53.5\% [47.0, 59.8] (123/230) \\
\midrule
DeepSeek-V3 & iid & 38.6\% [25.7, 53.4] (17/44) & 33.3\% [21.4, 48.0] (15/45) \\
DeepSeek-V3 & ood-host & 27.9\% [21.1, 35.9] (38/136) & 44.5\% [36.8, 52.5] (65/146) \\
DeepSeek-V3 & ood-property & 48.2\% [42.4, 54.1] (135/280) & 53.5\% [47.0, 59.8] (123/230) \\
\midrule
Qwen3-235B & iid & 40.9\% [27.7, 55.6] (18/44) & 46.7\% [32.9, 60.9] (21/45) \\
Qwen3-235B & ood-host & 36.8\% [29.1, 45.1] (50/136) & 53.4\% [45.3, 61.3] (78/146) \\
Qwen3-235B & ood-property & 55.0\% [49.1, 60.7] (154/280) & 51.3\% [44.9, 57.7] (118/230) \\
\bottomrule
\end{tabular}
\end{table}

\begin{table}[t]
\centering
\caption{Zero-shot logical-question accuracy on OOD-host by host structure family.}
\label{tab:zero-shot-by-family}
\begin{tabular}{lccccc}
\toprule
Family & $N_{\text{logical}}$ & Qwen2.5-72B & Llama 3.3-70B & DeepSeek-V3 & Qwen3-235B \\
\midrule
Garnet & 171 & 33.9\% (58/171) & 36.3\% (62/171) & 33.3\% (57/171) & 43.3\% (74/171) \\
Halide perov. & 83 & 27.7\% (23/83) & 30.1\% (25/83) & 38.6\% (32/83) & 48.2\% (40/83) \\
NASICON & 28 & 46.4\% (13/28) & 42.9\% (12/28) & 50.0\% (14/28) & 50.0\% (14/28) \\
\midrule
Total & 282 & 33.3\% (94/282) & 35.1\% (99/282) & 36.5\% (103/282) & 45.4\% (128/282) \\
\bottomrule
\end{tabular}
\end{table}

All four zero-shot baselines show reverse $>$ forward on OOD-host (Table~\ref{tab:zero-shot-by-direction}), with margins of 13--18pp for the 70B models and +16.6pp for Qwen3-235B, suggesting this asymmetry reflects task structure rather than model-specific bias. On the per-property axis (Table~\ref{tab:zero-shot-by-property}), all four baselines achieve higher band-gap than formation-energy accuracy on IID --- the opposite of the GRPO pattern (71.2\% FE vs.\ 47.8\% BG). This inversion suggests that zero-shot models rely on different heuristics than GRPO, which has learned ionic-radius and charge-balance reasoning that disproportionately benefits formation-energy questions.

\section{Site-Assignment Rules}
\label{app:site_rules}

\begin{table}[H]
\centering
\caption{Coordination-number rules for site assignment. For each family, cations are assigned to crystallographic sites based on their coordination number (CN) as computed by ChemEnv. Families marked with $\dagger$ use ionic-radius sorting instead of or in addition to CN rules.}
\label{tab:site_rules}
\small
\begin{tabular}{@{}llll@{}}
\toprule
Family & Site & CN rule & Notes \\
\midrule
Perovskite     & A-site & CN $\geq$ 8          & cuboctahedral \\
               & B-site & CN $\in \{5,6,7\}$   & octahedral \\
\addlinespace
Spinel         & A-site & CN $= 4$              & tetrahedral \\
               & B-site & CN $\in \{5,6,7\}$   & octahedral \\
\addlinespace
Olivine        & M-site & CN $\geq$ 5          & M1/M2 merged \\
               & T-site & CN $= 4$              & tetrahedral \\
\addlinespace
Scheelite      & A-site & CN $\geq$ 8          & \\
               & B-site & CN $\in \{4,5,6,7\}$ & \\
\addlinespace
Pyrochlore     & A-site & CN $\geq$ 8          & 8-coordinate \\
               & B-site & CN $\in \{5,6,7\}$   & octahedral \\
\addlinespace
Brownmillerite & A-site & CN $\geq$ 8          & \\
               & B-site & CN $\in \{4,5,6,7\}$ & oct.\ + tet.\ layers \\
\addlinespace
Garnet         & A-site & CN $\geq$ 8          & dodecahedral \\
               & B-site & CN $\in \{5,6,7\}$   & octahedral \\
               & C-site & CN $= 4$              & tetrahedral \\
\addlinespace
Halide perov.  & A-site & CN $\geq$ 8          & \\
               & B-site & CN $\in \{5,6,7\}$   & \\
\addlinespace
NASICON        & A-site & CN $\geq$ 8          & channel cation \\
               & B-site & CN $\in \{5,6,7\}$   & octahedral \\
               & C-site & CN $= 4$              & tetrahedral \\
\midrule
Layered oxide$^\dagger$       & \multicolumn{3}{l}{Sorted by Shannon ionic radius (larger $\to$ A-site)} \\
Double perov.$^\dagger$ & \multicolumn{3}{l}{A-site: CN $\geq$ 8;\; B/B$'$: sorted by ionic radius (larger $\to$ B$'$)} \\
\bottomrule
\end{tabular}
\end{table}

\section{Question Generation Details}
\label{app:question_gen}

\paragraph{Template construction.}
For each validated material at each substitutable site, a template composition is constructed by removing the element at the substituted site from the formula. Materials sharing the same (family, site, oxidation state, template) are grouped. For example, \ce{CaTiO3} and \ce{BaTiO3} at the A-site share the template \texttt{?-O3-Ti1} and belong to the same group. Within each group, all candidates share the same host framework and differ only at the substituted site.

\paragraph{Candidate selection.}
For each template group with $N \geq 3$ distinct elements, questions are generated as follows:
\begin{enumerate}
    \item Elements are ranked by the target property value (forward: best first; reverse: worst first).
    \item The gap filter is applied: the difference between rank 1 and rank 2 must exceed $\delta_{\min}$ (0.02~eV/atom for formation energy, 0.10~eV for band gap).
    \item For four-choice questions ($N \geq 4$): the correct answer (rank 1) and runner-up (rank 2) are fixed; two distractors are drawn from the remaining $N - 2$ elements. All $\binom{N-2}{2}$ subsets are generated without cap.
    \item For three-choice questions ($N = 3$): all three elements are included.
    \item Each unique (template group, candidate subset, property, direction) appears exactly once.
\end{enumerate}

\paragraph{Goal phrasing.}
Forward and reverse goals use distinct natural-language phrasings:

\begin{table}[h]
\centering
\small
\begin{tabular}{lll}
\toprule
Property & Forward goal & Reverse goal \\
\midrule
Formation energy & ``maximize thermodynamic stability'' & ``achieve the highest (least stable) \\
                 & & formation energy per atom'' \\
Band gap & ``maximize the electronic band gap'' & ``minimize the electronic band gap'' \\
\bottomrule
\end{tabular}
\end{table}

\section{SFT Trace Generation}
\label{app:sft_traces}

\paragraph{Prompt design.}
Reasoning traces are generated by prompting DeepSeek-R1~\citep{deepseek2025r1} with each training question and its correct answer. The model is instructed to reason as if arriving at the answer from scratch, without revealing that the answer was provided. The full prompts are shown below.

\begin{promptbox}
\textbf{System prompt}\\[4pt]
{\small\ttfamily
You are an expert materials scientist. You will be given a multiple-choice question about ionic substitution in crystalline materials. The correct answer is provided to you privately, but you must NOT reveal it upfront.\\[3pt]
Write your response as if you are reasoning through the problem from scratch and arriving at the answer through chemical logic. Do NOT say ``the correct answer is'' or ``I am told the answer is'' --- instead, analyze each candidate and build toward the conclusion naturally.\\[3pt]
Consider these factors:\\
- Ionic radius matching and site compatibility\\
- Coordination preferences and crystal field effects\\
- Tolerance factor and structural stability\\
- Electronegativity and bonding character\\
- Periodic trends and known material families\\[3pt]
Do NOT reference specific numerical property values (no eV, eV/atom, or band gap numbers). Reason ONLY from chemical principles and periodic table relationships.\\[3pt]
Evaluate each candidate, explain why the best one is superior and why each alternative is worse. Keep your reasoning concise (under 300 words). End with ``Answer: (X)'' where X is the letter you arrived at.}
\end{promptbox}

\begin{promptbox}
\textbf{System prompt}\\[4pt]
\textbf{User prompt} (example)\\[4pt]
{\small\ttfamily
Host material: LaAlO3 (perovskite)\\
Site to substitute: B-site\\
Design goal: ``maximize thermodynamic stability''\\[3pt]
Candidates:\\
\quad (A) Cu $\to$ LaCuO3\\
\quad (B) Co $\to$ LaCoO3\\
\quad (C) Al $\to$ LaAlO3\\
\quad (D) Cr $\to$ LaCrO3\\[3pt]
Which substitution best achieves the design goal? Explain your reasoning, then give your final answer as ``Answer: (X)'' where X is A, B, C, D.\\[3pt]
{[}The correct answer is (C) Al $\to$ LaAlO3. Explain why a materials scientist would arrive at this answer using chemical reasoning. Do not mention that you were told the answer.{]}}
\end{promptbox}

The bracketed instruction at the end is included only during trace generation and is stripped from the SFT training pairs. At training and evaluation time, only the question text appears in the user message.

\paragraph{Post-processing.}
Three steps are applied to each raw response:
\begin{enumerate}[itemsep=2pt,parsep=0pt,topsep=4pt]
    \item \textbf{Strip deliberation blocks:} Internal \texttt{<think>...</think>} blocks (DeepSeek-R1's chain-of-thought) are removed via regex, retaining only the final reasoning.
    \item \textbf{Remove leaked values:} Three regex patterns catch DFT value leaks:
    \begin{itemize}[itemsep=1pt,parsep=0pt,topsep=2pt]
        \item \texttt{-?\textbackslash d+\textbackslash.\textbackslash d+\textbackslash s*eV(?:/atom)?} (e.g., ``$-3.556$~eV/atom'')
        \item \texttt{(?:formation energy|band gap|energy)\textbackslash s*(?:of|=|:)\textbackslash s*-?\textbackslash d+\textbackslash.\textbackslash d+}
        \item \texttt{-?\textbackslash d+\textbackslash.\textbackslash d\{3,\}\textbackslash s*(?:eV|meV|Ry|Ha)} (high-precision values)
    \end{itemize}
    Matches are replaced with \texttt{[value removed]}.
    \item \textbf{Answer verification:} The final ``Answer: (X)'' is extracted and compared to ground truth. Mismatches are discarded.
\end{enumerate}
\paragraph{Results.}
Of 7,287 traces generated, 7,130 (97.8\%) pass all checks. Parse failures account for 144 (2.0\%), and 128 (1.8\%) contain leaked values that are cleaned to zero residual leaks. The median trace length is 1,837 characters. Yield is consistent across question types: forward (98.4\%) and reverse (97.4\%), formation energy (97.3\%) and band gap (99.7\%), three-choice and four-choice (both 97.8\%).

\section{Example GRPO Reasoning Traces}
\label{app:traces}

These traces show GRPO's reasoning on the same template group in both goal directions. The model correctly identifies $d$-electron occupancy as the primary driver of band-gap variation and reverses its conclusion when the goal flips. Its mechanistic framing occasionally overstates the instability of well-known compositions --- for example, the reverse trace describes Ni$^{3+}$ and Co$^{3+}$ as unstable in oxide perovskites, when LaNiO$_3$ and LaCoO$_3$ are well-known stable phases. The qualitative rankings are nonetheless correct, suggesting that GRPO learns reliable chemical intuition even when its stated justifications are imprecise.
\subsection*{Forward goal: maximize band gap.}

\noindent \textbf{Host:} TmAlO$_3$ (perovskite) \quad
\textbf{Site:} B-site \quad
\textbf{Goal:} maximize the electronic band gap \\
\textbf{Candidates:} (A) Al $\to$ TmAlO$_3$ \quad
(B) Ga $\to$ TmGaO$_3$ \quad
(C) In $\to$ TmInO$_3$ \quad
(D) Mn $\to$ TmMnO$_3$ \\
\textbf{Correct answer:} (A) Al $\to$ TmAlO$_3$

\begin{tcolorbox}[breakable, colback=blue!5, colframe=blue!60, title=GRPO model response (correct)]
The goal is to increase the band gap of TmAlO$_3$ by substituting the B-site cation. Band gap widening in oxides is favored by cations with high charge density and no partially filled $d$- or $f$-electrons, as these minimize electronic transitions. TmAlO$_3$ has a high band gap due to Al$^{3+}$ (no $d/f$ electrons) and Tm$^{3+}$ (localized $f$-electrons with no overlap for band formation). Evaluating the candidates:

\begin{itemize}
    \item \textbf{(A) Al $\to$ TmAlO$_3$:} This is identical to the host, retaining Al$^{3+}$ with high charge density and no $d/f$ electrons. No change occurs, maintaining the high band gap.
    \item \textbf{(B) Ga $\to$ TmGaO$_3$:} Ga$^{3+}$ has a larger ionic radius than Al$^{3+}$, reducing charge density. While still stable, Ga's higher polarizability may introduce minor electronic states, slightly lowering the band gap compared to Al.
    \item \textbf{(C) In $\to$ TmInO$_3$:} In$^{3+}$ has the largest ionic radius, further reducing charge density. This may increase polarizability and introduce more diffuse orbitals, potentially narrowing the band gap relative to Al.
    \item \textbf{(D) Mn $\to$ TmMnO$_3$:} Mn$^{3+}$ (if stable) has partially filled $d$-electrons, enabling Jahn--Teller distortions and mid-gap states, significantly reducing the band gap. Even Mn$^{2+}$ (if present) introduces low-energy transitions, making this unsuitable.
\end{itemize}

Candidate (A) preserves the optimal Al$^{3+}$ cation, ensuring no destabilizing electronic states or reduced charge density. Alternatives (B)--(D) introduce larger cations or $d$-electrons, risking band gap reduction. Thus, (A) best achieves the goal.

\vspace{0.3em}
\textbf{Answer: (A)}
\end{tcolorbox}

\vspace{0.5em}

\subsection*{Reverse goal: minimize band gap (same template group).}

\noindent \textbf{Host:} TmAlO$_3$ (perovskite) \quad
\textbf{Site:} B-site \quad
\textbf{Goal:} minimize the electronic band gap \\
\textbf{Candidates:} (A) Ga $\to$ TmGaO$_3$ \quad
(B) Ni $\to$ TmNiO$_3$ \quad
(C) In $\to$ TmInO$_3$ \quad
(D) Co $\to$ TmCoO$_3$ \\
\textbf{Correct answer:} (B) Ni $\to$ TmNiO$_3$

\begin{tcolorbox}[breakable, colback=blue!5, colframe=blue!60, title=GRPO model response (correct)]
The goal is to minimize the band gap in TmAlO$_3$ by substituting the B-site cation. Band gap reduction in perovskites is often achieved through introducing cations with partially filled $d$-orbitals that enable charge transfer transitions, which lower the energy required for electronic excitation. Among the candidates, Ni, Co, and Ga/In are transition metals, while Ga and In are post-transition metals with no $d$-electrons.

\begin{itemize}
    \item \textbf{Candidate (A) Ga:} Ga$^{3+}$ has no $d$-electrons, so TmGaO$_3$ remains a typical ionic perovskite with a large band gap. Although Ga is stable, it lacks the electronic structure for band gap reduction.
    \item \textbf{Candidate (B) Ni:} Ni$^{3+}$ (if stable) has $d$-electrons that can facilitate low-energy charge transfer transitions, potentially reducing the band gap. However, Ni$^{3+}$ is unstable in oxides, often decomposing to NiO, which may not form a stable perovskite. If formed, Ni-based perovskites like TmNiO$_3$ could exhibit smaller band gaps due to Ni's inherent electronic activity.
    \item \textbf{Candidate (C) In:} In$^{3+}$ also lacks $d$-electrons, similar to Ga. TmInO$_3$ would maintain a large band gap, as In-based perovskites are typically insulating without electronic transitions.
    \item \textbf{Candidate (D) Co:} Co$^{3+}$ has $d$-electrons that can reduce the band gap, but Co$^{3+}$ is unstable in oxides, prone to reduction to Co$^{2+}$ or CoO. TmCoO$_3$ may form, but Co's instability compared to Ni could limit band gap reduction effectiveness.
\end{itemize}

Ni is superior among the transition metals due to its higher tendency for charge transfer transitions in oxides, even if stability is a concern. Co is less favorable due to instability, while Ga and In are inert. Thus, Ni substitution best achieves the smallest band gap.
\vspace{0.3em}
\textbf{Answer: (B)}
\end{tcolorbox}

\vspace{0.5em}

\section{Training Hyperparameters}
\label{app:hyper}

\begin{table}[H]
\centering
\caption{Training hyperparameters for the SFT and GRPO stages. Both stages use LoRA adapters and are trained via the Tinker API.}
\label{tab:hyperparams}
\small
\begin{tabular}{@{}lcc@{}}
\toprule
Parameter & SFT & GRPO \\
\midrule
Base model              & Qwen3-8B            & SFT checkpoint \\
LoRA rank               & 32                  & 32 \\
LoRA alpha              & 32                  & 32 \\
LoRA target modules     & \multicolumn{2}{c}{all linear layers} \\
LoRA dropout            & 0.0                 & 0.0 \\
Learning rate           & 2e-4 (linear decay) & 4e-5 \\
Batch size              & 64                  & 64 questions $\times$ 8 rollouts \\
Epochs / steps          & 1 epoch (100 steps)  & 1 epoch (113 steps) \\
Completions per question & ---                & 8 \\
Max completion tokens   & 8{,}192             & 1{,}024 \\
Temperature             & ---                 & 0.8 \\
KL penalty              & ---                 & 0.0 (implicit via IS) \\
Loss function           & Cross-entropy       & Importance sampling \\
Constant-reward removal & ---                 & True \\
Optimizer               & Adam ($\beta_1\!=\!0.9$, $\beta_2\!=\!0.95$) & Adam \\
Training duration       & $\sim$5 min         & $\sim$2.5 hr \\
Compute cost            & $\sim$\$5           & $\sim$\$35 \\
\midrule
Trace generation cost   & \multicolumn{2}{c}{$\sim$\$10} \\
Total pipeline cost     & \multicolumn{2}{c}{$\sim$\$50} \\
\bottomrule
\end{tabular}
\end{table}

\begin{figure}[H]
  \centering
  \includegraphics[width=0.9\columnwidth]{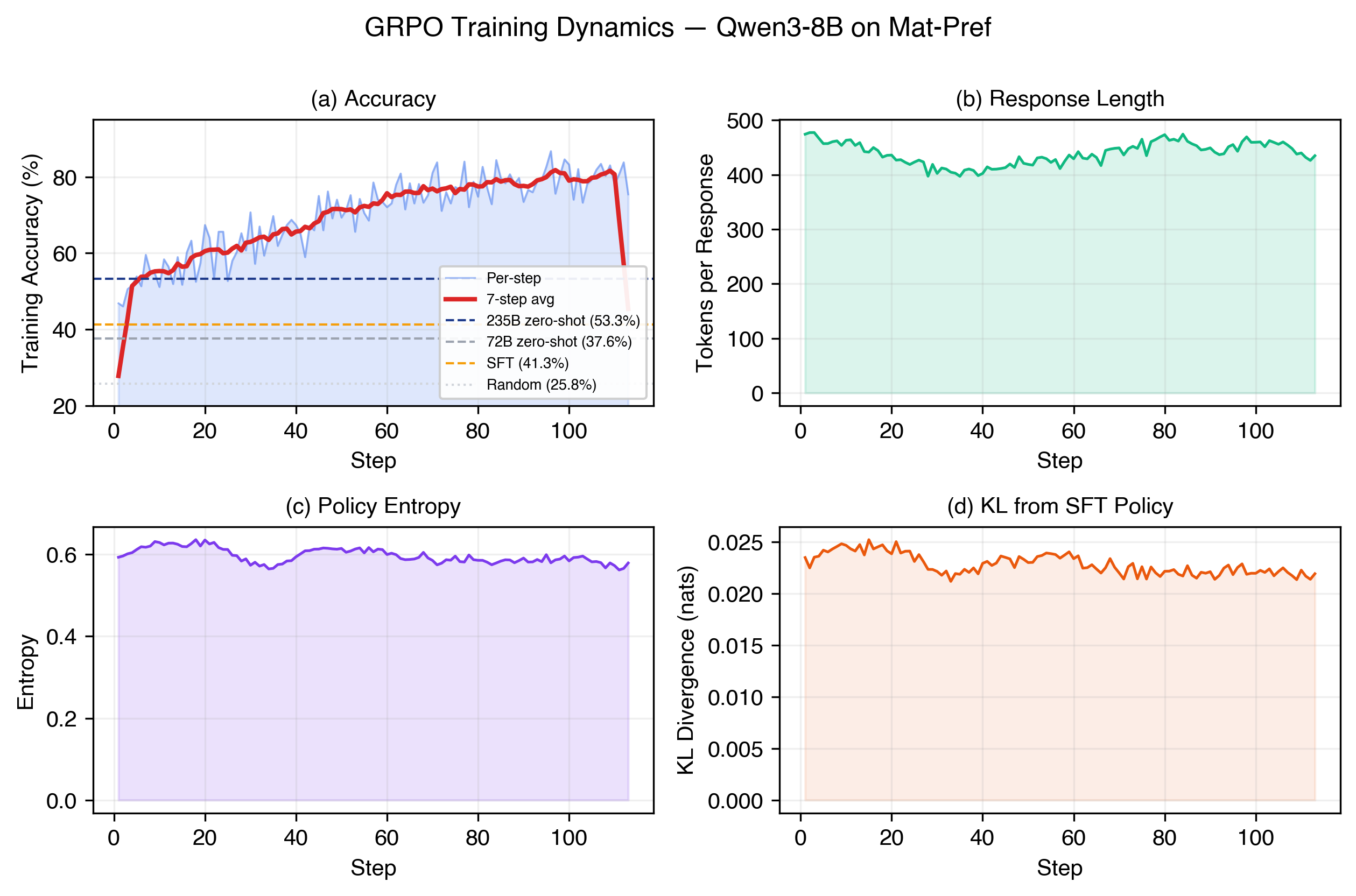}
  \caption{Full GRPO training dynamics (4-panel).
  (a)~Training accuracy. (b)~Mean response length remains stable
  at ${\sim}450$ tokens. (c)~Policy entropy decreases from 0.59
  to 0.58. (d)~KL divergence from SFT reference. Training uses
  batch size 64, group size 8, temperature 0.8, and LoRA rank 32.
  The total pipeline cost (trace generation + SFT + GRPO) is
  approximately \$50.}
  \label{fig:dynamics_full}
\end{figure}

\section{Confusion Analysis}
\label{app:confusion}

Raw error counts can be misleading because a single failed logical question generates errors across all of its distractor permutations. For example, Al→Ho appears 52 times in raw counts but derives from a single logical question. We therefore rank confusion pairs by the number of distinct logical questions they affect, and separate pairs driven by only 1--2 logical questions (\Cref{tab:confusion_artifacts}) from genuinely systematic confusions (\Cref{tab:confusion_lq}).

\begin{table}[H]
\centering
\caption{Top 15 confusion pairs ranked by distinct logical questions 
under the corrected grouping. ``Raw'' is the total instance count 
including distractor permutations; ``\#Logical'' counts distinct 
(template group, property, direction, site) tuples contributing to 
the pair.}
\label{tab:confusion_lq}
\small
\begin{tabular}{@{}llrrll@{}}
\toprule
Correct $\to$ Predicted & \#Logical & \#Templates & Raw & Property & \#Families \\
\midrule
Cu $\to$ Co  & 8 & 7 & 15 & band gap   & 5 \\
Fe $\to$ Cr  & 8 & 8 & 14 & mixed      & 5 \\
Li $\to$ K   & 7 & 6 &  7 & mixed      & 4 \\
K $\to$ Li   & 7 & 5 &  7 & mixed      & 2 \\
Lu $\to$ La  & 6 & 6 & 18 & band gap   & 4 \\
Tm $\to$ Sc  & 6 & 6 & 11 & band gap   & 2 \\
Cr $\to$ In  & 6 & 4 &  9 & form.\ en. & 1 \\
Cr $\to$ Bi  & 6 & 4 &  8 & mixed      & 3 \\
Ti $\to$ V   & 6 & 6 &  7 & band gap   & 3 \\
Pd $\to$ Hg  & 5 & 4 & 17 & mixed      & 1 \\
Cu $\to$ Ni  & 5 & 4 & 12 & mixed      & 4 \\
Cr $\to$ Fe  & 5 & 5 & 12 & mixed      & 3 \\
Ca $\to$ Ba  & 5 & 5 & 10 & band gap   & 3 \\
Co $\to$ Bi  & 5 & 2 &  9 & mixed      & 1 \\
Sb $\to$ Ta  & 5 & 5 &  7 & band gap   & 2 \\
\bottomrule
\end{tabular}
\end{table}

The top confusion pairs recur across 4--5 structure families, confirming they reflect genuine chemical blind spots rather than template-specific failures. The most common are first-row transition-metal swaps: Cu$\leftrightarrow$Co (8 logical questions, 5 families) and Fe$\to$Cr (8 logical questions, 5 families), where similar ionic radii among 3$d$ cations make substitutions difficult to distinguish. Lu$\to$La (6 logical questions, 4 families) reflects lanthanide-series confusion, where the lanthanide contraction produces chemically similar cations. Li$\leftrightarrow$K (7+7 logical questions) is the only alkali-metal confusion in the top 15, suggesting the model occasionally reverses size-based rankings despite size-and-charge reasoning being its strongest capability.

\begin{table}[H]
\centering
\caption{High-raw-count confusion pairs that derive from only 1--2 logical questions. These eight pairs account for $\sim$250 raw errors but only 10 distinct logical questions, and are excluded from the main confusion analysis (\Cref{tab:confusion_lq}).}
\label{tab:confusion_artifacts}
\small
\begin{tabular}{@{}lrrll@{}}
\toprule
Pair & Raw & \#LQ & Family & Property \\
\midrule
Al $\to$ Ho  & 52 & 1 & Layered oxide & Band gap \\
B $\to$ Lu   & 42 & 1 & Spinel & Band gap \\
Th $\to$ Zr  & 38 & 2 & Perovskite & Mixed \\
Pd $\to$ Pt  & 31 & 1 & Perovskite & Form.\ en. \\
In $\to$ La  & 30 & 2 & Double perov./Pyrochlore & Band gap \\
Ce $\to$ Cr  & 22 & 1 & Spinel & Band gap \\
Eu $\to$ Fe  & 18 & 1 & Scheelite & Band gap \\
B $\to$ Er   & 17 & 1 & Layered oxide & Band gap \\
\bottomrule
\end{tabular}
\end{table}

\section{SFT $\to$ GRPO Logical-Question Flow}
\label{app:question_flow}

\begin{table}[H]
\centering
\caption{SFT $\to$ GRPO transition matrix on logical questions, per test split. Cells show joint outcomes (both correct, SFT-only correct, GRPO-only correct, both wrong); the GRPO-only minus SFT-only column gives the net gain in logical questions.}
\label{tab:sft-grpo-transitions}
\begin{tabular}{lcccccc}
\toprule
Split & $N_{\text{logical}}$ & Both \checkmark & SFT \checkmark only & GRPO \checkmark only & Both \xmark & $\Delta_{\text{net}}$ \\
\midrule
IID          &  89 &  37 &  3 &  21 &  28 & +18 \\
OOD-host     & 282 & 117 & 18 &  85 &  62 & +67 \\
OOD-property & 510 & 199 & 57 & 107 & 147 & +50 \\
\midrule
Total        & 881 & 353 & 78 & 213 & 237 & +135 \\
\bottomrule
\end{tabular}
\end{table}

The fix-to-regress ratio is highest on IID (21 fixes vs.\ 3 regressions, 7.0$\times$), drops to 4.72$\times$ on OOD-host, and reaches 1.88$\times$ on OOD-property where band-gap-only composition is harder for both models.

\begin{table}[H]
\centering
\caption{Per-family SFT $\to$ GRPO transitions on the OOD-host split.}
\label{tab:sft_grpo_family}
\small
\begin{tabular}{@{}lcccccc@{}}
\toprule
Family & $N_{\text{logical}}$ & Both \checkmark & SFT \checkmark only & GRPO \checkmark only & Both \xmark & $\Delta_{\text{net}}$ \\
\midrule
Garnet            & 171 &  65 & 10 & 57 & 39 & +47 \\
Halide perovskite &  83 &  34 &  6 & 23 & 20 & +17 \\
NASICON           &  28 &  18 &  2 &  5 &  3 &  +3 \\
\midrule
Total             & 282 & 117 & 18 & 85 & 62 & +67 \\
\bottomrule
\end{tabular}
\end{table}

All three OOD-host families show net gains from GRPO. Garnet (+47) and halide perovskite (+17) contribute the bulk of the OOD-host improvement; NASICON's smaller +3 net reflects its small base ($n = 28$) and the fact that SFT already solved a large fraction (18/28 both-correct) before GRPO. GRPO's 85 recoveries are not biased toward either property (47\% BG vs 46\% baseline), indicating that GRPO improves both properties roughly proportionally on OOD-host.

\section{Consistency Analysis}
\label{app:consistency}

\begin{table}[H]
\centering
\caption{Logical-question accuracy under three aggregation rules for SFT and GRPO. The lenient--strict gap quantifies within-LQ permutation sensitivity (lower = more consistent). Within-LQ rate is the mean per-LQ fraction of correct permutations. CIs are cluster bootstrap 95\% over template groups ($B = 10{,}000$).}
\label{tab:consistency}
\small
\setlength{\tabcolsep}{3pt}
\begin{tabular}{@{}llcccccc@{}}
\toprule
Model & Split & $N$ & Lenient & Majority & Strict & Lenient$-$Strict & Within-LQ rate \\
\midrule
SFT  & IID          &  89 & 61.8\% & 44.9\% & 33.7\% & 28.0pp [14.1, 43.7] & 46.7\% [38.2, 55.1] \\
SFT  & OOD-host     & 282 & 62.4\% & 47.9\% & 41.5\% & 20.9pp [13.8, 28.6] & 51.0\% [45.0, 57.1] \\
SFT  & OOD-property & 510 & 63.9\% & 50.2\% & 40.8\% & 23.1pp [18.4, 28.0] & 51.7\% [47.5, 55.7] \\
\midrule
GRPO & IID          &  89 & 69.7\% & 65.2\% & 56.2\% & 13.5pp [4.9, 23.2]  & 64.5\% [54.1, 75.0] \\
GRPO & OOD-host     & 282 & 77.7\% & 71.6\% & 63.8\% & 13.8pp [8.5, 19.7]  & 71.4\% [65.7, 76.8] \\
GRPO & OOD-property & 510 & 67.3\% & 60.0\% & 51.6\% & 15.7pp [12.0, 19.6] & 60.0\% [55.9, 64.1] \\
\bottomrule
\end{tabular}
\end{table}

\section{Logit Lens Analysis}
\label{app:logit_lens}

We apply the logit lens~\citep{belrose2023tuned} to compare how SFT and GRPO form answer decisions internally. At each layer, we project the residual stream at the last prompt token (before any generation) through the final RMSNorm and unembedding matrix and measure the \emph{crystallization rate}: the fraction of questions where the argmax over answer-letter logits matches the model's final output.

\begin{figure*}[t]
\centering
\includegraphics[width=\columnwidth]{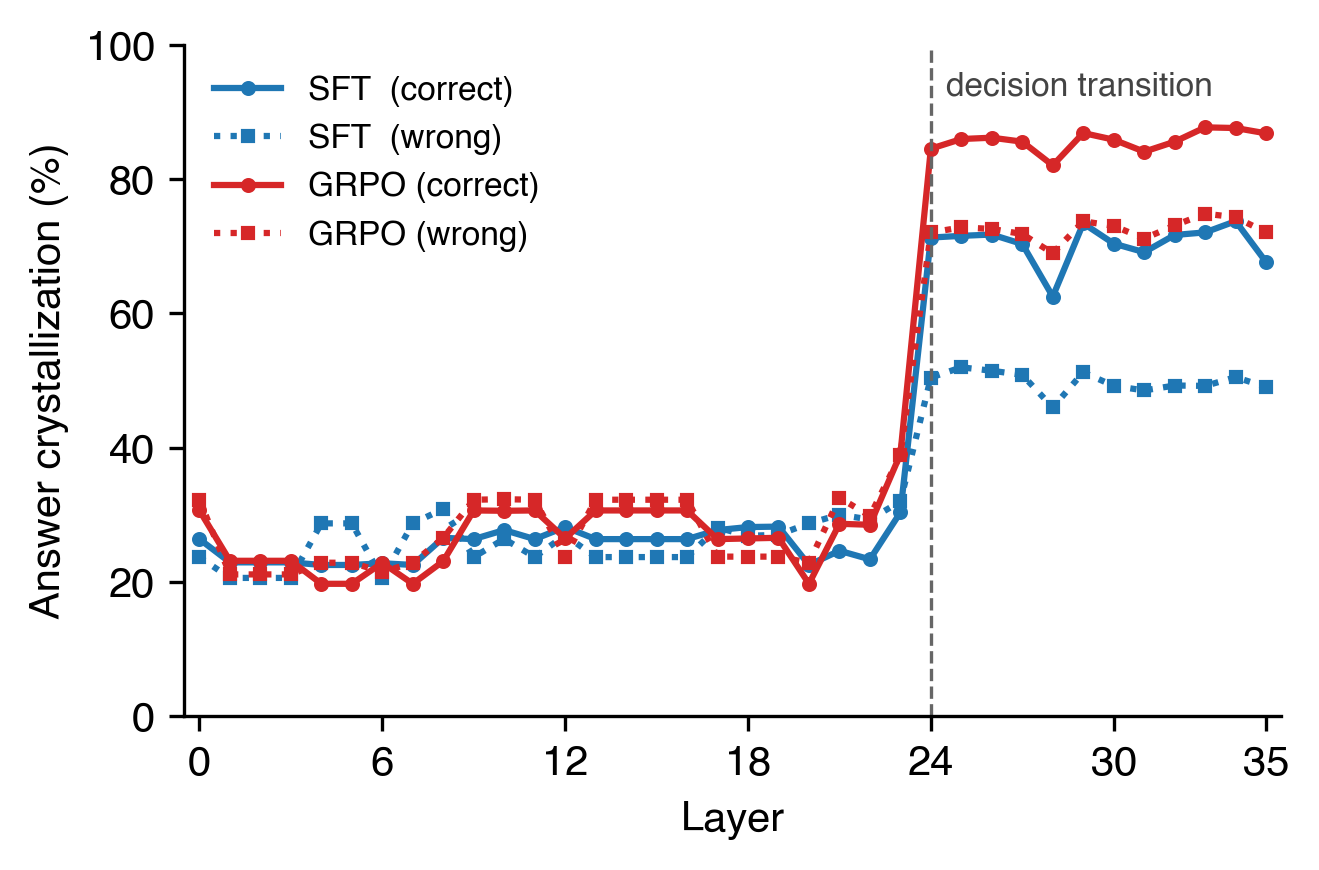}
\caption{Logit-lens crystallization rate vs.\ layer. Both models transition from random ($\sim$22\%) to committed at layer 24; GRPO sustains a $\sim$20pp advantage thereafter.}
\label{fig:logit-lens}
\end{figure*}

Both checkpoints transition sharply from near-random ($\sim$22\%) to committed at layer 24, but GRPO crystallizes at 79.6\% vs.\ SFT's 60.2\%, sustaining a $\sim$20pp advantage through the final layer (\Cref{fig:logit-lens}). GRPO sharpens commitment at the same decision stage rather than deciding earlier. Linear probes on the same activations confirm that input-level
features are encoded identically, but GRPO's residual stream is
better at predicting its own correctness. We train logistic
regression probes (L2, $C{=}1.0$) on the 4,096-dimensional
post-block residual stream at the last prompt token, using a
stratified 80/20 split of 3,334 test questions at each layer.
Probes for target property, goal direction, and structure family
reach 100\% accuracy for both checkpoints, confirming input
features are linearly accessible regardless of training stage.
On binary correctness prediction, GRPO peaks at 73.0\% (layer 21)
vs.\ SFT's 66.4\% (layer 19), a +6.6pp gap, with the maximum
per-layer gap of +12.1pp at layer 15.

\section{Trace Length and Accuracy}
\label{app:trace_length}

Incorrect traces are modestly longer than correct ones (1,764 vs. 1,712 characters), and accuracy degrades with trace length (\Cref{tab:trace_acc_new}), though we cannot distinguish confident reasoning from easier questions as the cause. The model never produces responses under 500 characters, confirming no length collapse.

\begin{table}[H]
\centering
\caption{Accuracy by trace length bin. Counts are raw question instances (not logical questions), since trace length varies across permutations of the same logical question.}
\label{tab:trace_acc_new}
\small
\begin{tabular}{@{}lcc@{}}
\toprule
Trace length (chars) & Accuracy & $n$ \\
\midrule
$[500, 1{,}000)$   & 71.2\% &    66 \\
$[1{,}000, 2{,}000)$ & 61.2\% & 2{,}704 \\
$[2{,}000, \infty)$  & 55.0\% &   564 \\
\bottomrule
\end{tabular}
\end{table}

\section{Halide Perovskite Details}
\label{app:halide_details}

\begin{table}[H]
\centering
\caption{GRPO logical-question accuracy on halide perovskite OOD-host split, broken down by halide anion.}
\label{tab:halide-by-anion}
\begin{tabular}{lcc}
\toprule
Anion & Accuracy & $N_{\text{logical}}$ \\
\midrule
F  & 71.7\% [57.5, 82.7] & 46 \\
Cl & 76.9\% [49.7, 91.8] & 13 \\
Br & 60.0\% [38.7, 78.1] & 20 \\
I  & 50.0\% [15.0, 85.0] &  4 \\
\bottomrule
\end{tabular}
\end{table}

\section{Question-Level vs.\ Logical-Question-Level Comparison}
\label{app:question_level}

For comparability with standard multiple-choice evaluation, \Cref{tab:ql_comparison} reports both question-level and logical-question-level accuracy for all (model, split) cells.

\begin{table}[H]
\centering
\caption{Question-level vs.\ logical-question-level accuracy, with Wilson 95\% CIs. Question-level treats each permutation independently; logical-level scores by majority vote across permutations.}
\label{tab:ql_comparison}
\small
\setlength{\tabcolsep}{3pt}
\begin{tabular}{@{}llll@{}}
\toprule
Model & Split & Question-level [95\% CI] & Logical-level [95\% CI] \\
\midrule
Qwen2.5-72B & IID          & 134/356 = 37.6\% [32.8, 42.8]         & 33/89 = 37.1\% [27.8, 47.5] \\
            & OOD-host     & 274/700 = 39.1\% [35.6, 42.8]         & 94/282 = 33.3\% [28.1, 39.0] \\
            & OOD-property & 954/2{,}278 = 41.9\% [39.9, 43.9]     & 202/510 = 39.6\% [35.5, 43.9] \\
\midrule
SFT         & IID          & 147/356 = 41.3\% [36.3, 46.5]         & 40/89 = 44.9\% [35.0, 55.3] \\
            & OOD-host     & 329/700 = 47.0\% [43.3, 50.7]         & 135/282 = 47.9\% [42.1, 53.7] \\
            & OOD-property & 1{,}082/2{,}278 = 47.5\% [45.5, 49.6] & 256/510 = 50.2\% [45.9, 54.5] \\
\midrule
GRPO        & IID          & 215/356 = 60.4\% [55.2, 65.3]         & 58/89 = 65.2\% [54.8, 74.3] \\
            & OOD-host     & 471/700 = 67.3\% [63.7, 70.7]         & 202/282 = 71.6\% [66.1, 76.6] \\
            & OOD-property & 1{,}327/2{,}278 = 58.3\% [56.2, 60.3] & 306/510 = 60.0\% [55.7, 64.2] \\
\bottomrule
\end{tabular}
\end{table}

For GRPO, logical-level accuracy exceeds question-level on IID and OOD-host, reflecting the within-question consistency discussed in \Cref{sec:consistency}. For the zero-shot baseline, the pattern reverses, confirming that majority voting penalizes inconsistent reasoning.

\section{External Benchmark Evaluation}
\label{app:external}

\begin{table}[H]
\centering
\caption{External benchmark accuracy (\%). SFT and GRPO are directly comparable under the same evaluation protocol. GRPO training on \matpref{} does not cause catastrophic forgetting.}
\small
\begin{tabular}{@{}lcc@{}}
\toprule
Benchmark & SFT & GRPO \\
\midrule
MMLU College Chemistry (100)  & 62.0\% & 59.0\% \\
MMLU HS Chemistry (203)       & 84.7\% & 86.2\% \\
ARC-Challenge (1{,}172)       & 90.6\% & 91.2\% \\
\bottomrule
\end{tabular}
\end{table}

Performance is stable across all benchmarks. The $-3$pp shift on college chemistry is within noise for $n = 100$. High-school chemistry and ARC-Challenge show marginal improvements, confirming that domain-specific GRPO training preserves general reasoning ability.

\section{HLE-Gold Chemistry Question IDs}

For reproducibility, Table~\ref{tab:hle-gold-ids-full} lists the 42 text-only chemistry question IDs from HLE-Gold Bio/Chem used in the external evaluation in Section~6. To avoid reproducing the benchmark content in full, we identify each question by ID and summarize only a few representative cases below. These examples are drawn from the complete 42-question run and are included only to illustrate the behavioral contrast between the base model and the GRPO-trained model.

\begin{table}[H]
\centering
\caption{Question IDs in the validated HLE-Gold chemistry evaluation set.}
\label{tab:hle-gold-ids-full}
\footnotesize
\setlength{\tabcolsep}{4pt}
\renewcommand{\arraystretch}{1.1}
\begin{tabular}{p{0.31\textwidth} p{0.31\textwidth} p{0.31\textwidth}}
\toprule
\texttt{66f87ab781a069162c8e7cd2} & \texttt{6716f035bab94116769c0082} & \texttt{6720cd0acf47ec0733864dd8} \\
\texttt{66fc5b54ffa390c4af01820f} & \texttt{6717ac23a5c8a6a9392b1b34} & \texttt{67228be893273f2ea4d39e36} \\
\texttt{66fc5e8f98a7264ef58309b9} & \texttt{671808958b88f01935b5825a} & \texttt{67230f05092b2c17f66c84aa} \\
\texttt{66fe16f4762ab58b937863b8} & \texttt{67182659782f2aff686eab73} & \texttt{6723a613f747d32c6b0b65dc} \\
\texttt{66fec7825e6051260840e060} & \texttt{671af1baed3d54e87368bca2} & \texttt{67254420046774791d0f1a2e} \\
\texttt{66fecbff69d5712b5401553e} & \texttt{671bbe5a85b315d9f07698a9} & \texttt{6725592b4ee96e5b68bb383b} \\
\texttt{66ff0a666dc4b2ab76a19db9} & \texttt{671c0ccd29d77939ea7f3bd4} & \texttt{67257157c16289d7e113915b} \\
\texttt{6700497e7c6e32185d473d3e} & \texttt{671c0d6f66154e5cf55994de} & \texttt{6725b2ba569cf0c62da64cd1} \\
\texttt{67008a05ad0fee7d7b4efb3c} & \texttt{671d91bcad7fb0793a0e93bd} & \texttt{672ba16ae227dfbd739cfcc3} \\
\texttt{6700a5f67899925724a8fa81} & \texttt{671db218fe1146e348ef1266} & \texttt{6730f3c006cd33fe46ca2dfe} \\
\texttt{670880520ed68fbdc467064e} & \texttt{671dba3e5102c27a58a6c501} & \texttt{673681def5487e4de6e78e1e} \\
\texttt{67089c013b511a575fc306de} & \texttt{671e7fd05cd705ffbd3faab7} & \texttt{6736acee69d6d88d06d3124a} \\
\texttt{670edc9dbddc0cfe673272c8} & \texttt{671e8c73745919354c759204} & \texttt{673cd3821cf7bd0b7cc91fe5} \\
\texttt{67151b015fc8ee8feaa3538c} & \texttt{671f33cb75523fe63c0a8b60} & \texttt{67833c10f5a1ca415084accd} \\
\bottomrule
\end{tabular}

\end{table}

\section{Representative Trace Comparisons on Validated HLE-Gold Chemistry}
\label{app:hle-gold-traces}

To illustrate the behavioral contrast discussed in Section~6, we report two representative examples from the full 42-question HLE-Gold chemistry evaluation. These are not a separate evaluation subset; they are qualitative illustrations drawn from the complete validated run. Unless otherwise noted, text in black reproduces the model output verbatim; blue bracketed text indicates editorial markers or omitted spans added for readability.

\paragraph{Example 1: both models are correct, but the trained trace is much shorter and more decisive.}
\textbf{HLE ID:} \texttt{670edc9dbddc0cfe673272c8} \\
\textbf{Gold answer:} \texttt{B} \\
\textbf{Base extracted final answer:} \texttt{B} \\
\textbf{Trained extracted final answer:} \texttt{B}

\begin{promptbox}
\small
\textbf{Prompt (verbatim).}

Would you please provide a comprehensive list of all possible organic A-site cations that are capable of independently forming three-dimensional lead halide perovskite structures (such as A-Pb-Br$_3$)?

Answer Choices: \\
A. Cesium, Methylammonium, Formamidinium \\
B. Cesium, Methylammonium, Formamidinium and Aziridinium \\
C. Cesium, Methylammonium, Formamidinium and Ethylammonium \\
D. Cesium, Methylammonium, Formamidinium and Methylhydrazinium \\
E. Cesium, Methylammonium, Formamidinium and Dimethylammonium
\end{promptbox}

\begin{tracebox}{Base Model Trace Excerpt (2469 words total)}
\small
\ttfamily
\textcolor{blue!60!black}{\textit{[raw model output begins]}}

Okay, so I need to figure out which organic A-site cations can form three-dimensional lead halide perovskites like A-Pb-Br3. The answer choices are A to E, each listing different cations. Let me start by recalling what I know about perovskite structures.

Perovskites have the general formula ABX3, where A is a cation, B is a metal (like Pb in this case), and X is a halide (like Br, Cl, I). The A-site cation is usually an organic or inorganic ion. For three-dimensional structures, the A cation needs to be large enough to fit in the perovskite lattice without causing distortion. If the A cation is too small, it might lead to a 2D or 1D structure instead of 3D.

The common organic A cations I remember are methylammonium (MA+), formamidinium (FA+), and cesium (Cs+). Cesium is inorganic, but it's often used as an A-site cation in perovskites. Wait, the question says "organic A-site cations," but cesium is inorganic. Hmm, maybe the question is a bit ambiguous. Let me check the answer choices again. The options include cesium in all of them, so maybe the question allows for inorganic cations as well. But the question specifically says "organic A-site cations." Wait, maybe the user made a mistake, or maybe they consider cesium as a possible A-site even though it's inorganic. Let me think.

In perovskites, the A-site can be either organic or inorganic. For example, in MAPbI3, MA+ is organic. CsPbI3 is inorganic. But the question says "organic A-site cations," so maybe the answer should exclude cesium? But the answer choices all include cesium. Wait, maybe the question is not strictly about organic, but the options are given with cesium included. Maybe the user is considering that cesium is a possible A-site even though it's inorganic. Let me check the options again.

Looking at the answer choices, all options include cesium. So perhaps the question is not strictly about organic, but the options are given with cesium as a possible answer. Maybe the question is a bit off, but I need to go with the given options.

\textcolor{blue!60!black}{\textit{[... excerpted for brevity ...]}}

In the case of dimethylammonium, the structure is (CH3)2NH2+. This cation is larger than MA+ and might fit into the lattice. Similarly, ethylammonium is C2H5NH3+, which is also larger. Both can form 3D perovskites. But which of these is included in the answer choices?

The options are:

C. Cs, MA, FA, and Ethylammonium

E. Cs, MA, FA, and Dimethylammonium

So, both are possible. But I think that the answer is E, because dimethylammonium is more commonly used in some studies. Alternatively, the answer is C. But I'm not sure. Alternatively, the answer is B, but I don't think aziridinium is used. Given that, I think the correct answer is C or E. But I need to check.

Wait, in the original question, the example is A-Pb-Br3.

In the case of Cs, it's CsPbBr3, which is a 3D perovskite.

MA is CH3NH3PbBr3, which is 3D.

FA is CH(NH2)2PbBr3, which is 3D.

Ethylammonium is C2H5NH3PbBr3, which is 3D.

Dimethylammonium is (CH3)2NH2PbBr3, which is 3D.

So, all of these can form 3D structures.

But the answer choices are given.

If the question is from a textbook, the answer might be C or E.

But I need to think about which of these is more commonly accepted.

In some references, the common A-site cations are MA, FA, Cs, and sometimes EA.

Dimethylammonium is also used, but maybe less common.

So, if the answer choices are given, and the options are C and E, then the answer might be C.

But I'm not sure.

Alternatively, the answer is B, but I don't think aziridinium is used.

Given that, I think the correct answer is C or E.

But since the options are given, and the user is asking for the answer, I need to choose between them.

In some sources, the answer is C: Cs, MA, FA, and Ethylammonium.

But I'm not certain.

Alternatively,
\end{tracebox}

\begin{tracebox}[colback=green!3,colframe=green!35!black,fonttitle=\bfseries]{Trained Model Full Output (486 words total)}
\small
\ttfamily
\textcolor{blue!60!black}{\textit{[raw model output begins]}}

The question asks for a comprehensive list of all possible organic A-site cations capable of independently forming three-dimensional lead halide perovskite structures, such as A-Pb-Br$_3$. To determine the correct answer, we must evaluate the structural and chemical requirements for an organic cation to form such perovskites.

First, the A-site cation in perovskites typically occupies the 12-coordinate site, requiring a large, symmetric, and flexible structure to accommodate the perovskite lattice without causing distortions. The cation must be relatively inert to avoid decomposition or instability in the perovskite phase. Additionally, the cation should not introduce significant steric hindrance or charge imbalance that could disrupt the perovskite structure.

Evaluating the answer choices:

- **Cesium (Cs$^+$)**: A monovalent inorganic cation with no organic groups. It is well-established for forming stable, three-dimensional perovskites like CsPbI$_3$ due to its large size and symmetric charge distribution, which minimizes strain in the lattice.

- **Methylammonium (CH$_3$NH$_3^+$)**: A common organic cation in perovskites (e.g., MAPbI$_3$). Its small size and asymmetric structure can lead to phase segregation, but it is still effective for forming 3D perovskites under certain conditions.

- **Formamidinium (CH(NH$_2$)$_2^+$)**: Similar to methylammonium, formamidinium is used in perovskites (e.g., FAPbI$_3$). Its larger size compared to methylammonium reduces strain, but it still forms 3D perovskites effectively.

- **Aziridinium**: This cation has a strained, cyclic structure with high charge density. While it can form perovskites, the strain may lead to instability or phase separation, making it less reliable for consistent 3D perovskite formation compared to the others.

- **Ethylammonium**: This cation is larger than methylammonium but still has asymmetric charge distribution. While it can form perovskites, its size may not be optimal for maintaining the symmetric A-site required for stable 3D structures.

- **Methylhydrazinium**: This cation has a high charge density and potential for instability due to its hydrazine-like structure. It is less commonly used in perovskites and may not form stable 3D perovskites without additional stabilizers.

- **Dimethylammonium**: This cation is larger than methylammonium but still asymmetric. While it can form perovskites, its size and charge distribution may not be ideal for maintaining the symmetric A-site required for stable 3D perovskites.

The key distinction lies in the inclusion of **aziridinium** in option B. While aziridinium can form perovskites, its strained structure and high charge density make it less reliable for consistently forming stable, three-dimensional perovskites compared to the other cations listed. The other options (Cesium, Methylammonium, Formamidinium) are well-documented for forming stable 3D perovskites, with no evidence supporting the inclusion of aziridinium as a primary candidate.

Thus, the most comprehensive and accurate answer is **B**, as it includes all the established cations (Cesium, Methylammonium, Formamidinium) and adds aziridinium, which, while less common, is still a valid cation for perovskite formation. However, the inclusion of aziridinium may be considered less optimal due to potential instability, but the question asks for all possible cations, not just the most reliable ones.

**Answer: B. Cesium, Methylammonium, Formamidinium and Aziridinium**
\end{tracebox}

\textit{Interpretation.} Both models eventually reach the correct answer, but the base trace remains unresolved and oscillatory almost to the end, whereas the trained trace adopts a clearer structure and commits decisively.

\vspace{0.8em}

\paragraph{Example 2: the trained model is much more decisive, but that decisiveness does not guarantee correctness.}
\textbf{HLE ID:} \texttt{671c0d6f66154e5cf55994de} \\
\textbf{Gold answer:} \texttt{D} \\
\textbf{Base extracted final answer:} \texttt{none of the options. But since that's not an option, maybe I'm misunderstanding the question.} \\
\textbf{Trained extracted final answer:} \texttt{E}

\begin{promptbox}
\small
\textbf{Prompt (verbatim).}

What achiral and non polar crystal classes have the correct symmetry for optical activity?

Answer Choices: \\
A. m and mm2 \\
B. -6, -62m, and -43m \\
C. 3m, 4m, and 6mm \\
D. -4 and -42m \\
E. 1, 2, 3, 4, and 6
\end{promptbox}

\begin{tracebox}{Base model trace excerpt (2975 words total)}
\small
\ttfamily
\textcolor{blue!60!black}{\textit{[raw model output begins]}}

Okay, so I need to figure out which achiral and non-polar crystal classes have the correct symmetry for optical activity. Let me start by recalling what optical activity means. Optical activity is the ability of a substance to rotate plane-polarized light, right? And this is typically associated with chiral molecules. But wait, the question is about crystal classes, not molecules. So maybe I need to think about the symmetry of the crystal itself.

Wait, but the question says "achiral and non-polar" crystal classes. So the crystal classes themselves are achiral and non-polar, but they still have the symmetry that allows for optical activity? That seems contradictory. Because if a crystal is achiral, it shouldn't have optical activity, right? Because optical activity is a chiral property. So maybe there's a misunderstanding here. Let me think again.

Wait, maybe the question is about the crystal's symmetry allowing for the possibility of optical activity, even though the crystal itself is achiral. But how? Because if the crystal is achiral, then it can't be optically active. Unless there's some other factor. Wait, maybe the crystal is not chiral, but the way it's structured allows for optical activity? Hmm, I'm confused.

Let me recall some crystallography. Optical activity in crystals is usually due to chirality. But if a crystal is achiral, meaning it has a center of inversion or a mirror plane, then it can't be optically active. So maybe the question is a trick question? But the answer choices are given, so I need to check which of them are achiral and non-polar, and then see if they can have optical activity.

\textcolor{blue!60!black}{\textit{[... excerpted for brevity ...]}}

Option B: -6, -62m, and -43m. These are crystal classes with a center of inversion. For example, the class -6 has a six-fold axis and a center of inversion. So, they are non-polar and achiral. But they are not optically active. So, they are not the answer.

Option C: 3m, 4m, and 6mm. These classes have mirror planes and possibly other symmetries. For example, 3m has a three-fold axis and mirror planes. They are achiral (since they have mirror planes) and non-polar? Wait, if they have a mirror plane, they might not have a center of inversion. So, they are polar. So, they are not non-polar. So, option C is not correct.

Option D: -4 and -42m. These classes have a center of inversion. For example, -4 has a four-fold axis and a center of inversion. So, they are non-polar and achiral. But again, they are not optically active. So, not the answer.

So, all options except E are achiral and non-polar? Wait, no. Let me check again.

Wait, the non-polar crystal classes are those with a center of inversion. So, the classes that have a center of inversion are the ones that are non-polar. So, for example, the class -4 has a center of inversion, so it's non-polar. The class m (option A) does not have a center of inversion, so it's polar. So, the achiral and non-polar crystal classes are those that have a center of inversion. So, the classes like -4, -42m, -6, -62m, -43m, etc. So, option B and D have some of these. But none of them are optically active, since they are achiral.

So, the question is asking for achiral and non-polar crystal classes that have the correct symmetry for optical activity. But if they are achiral, they can't be optically active. So, the answer is none of the options. But since that's not an option, maybe the question is wrong, or I'm missing something.

Wait, maybe the question is asking about the crystal's symmetry allowing for the possibility of optical activity in some way, even though the crystal is achiral. For example, if the crystal has a screw axis, but that's not directly related. Or maybe the question is referring to the fact that some achiral crystals can have optical activity
\end{tracebox}

\begin{tracebox}[colback=green!3,colframe=green!35!black,fonttitle=\bfseries]{Trained Model Full Output (530 words total)}
\small

\ttfamily
\textcolor{blue!60!black}{\textit{[raw model output begins]}}

The question asks which achiral and non-polar crystal classes have the correct symmetry for optical activity. To determine this, we need to recall that optical activity requires a lack of a center of symmetry and the presence of a single enantiomer, which is only possible in chiral molecules. However, the crystal class itself must not be chiral, meaning it must not have any symmetry elements that would make it equivalent to a chiral object. Additionally, the crystal must be non-polar, so it cannot have a polar axis or other polarizing features.

First, we consider the symmetry elements of each crystal class:

- **A. m and mm2**: These classes have mirror planes and dihedral symmetry. The presence of mirror planes indicates chirality is not possible, as mirror images would be equivalent. However, these classes are not inherently achiral or non-polar; they can be part of polar crystal systems.

- **B. -6, -62m, and -43m**: These classes include improper rotations. The presence of these symmetries can sometimes allow for optical activity, but these classes are not typically considered the relevant answer here.

- **C. 3m, 4m, and 6mm**: These classes have high symmetry, including mirror planes and rotational axes. The presence of mirror planes may not allow for optical activity, as chirality is not maintained.

- **D. -4 and -42m**: These classes include improper rotations, but the model rejects them as not the best fit.

- **E. 1, 2, 3, 4, and 6**: These are described as low-symmetry crystal classes with no mirror planes or centers of symmetry.

These classes are achiral because they lack mirror planes or centers of symmetry, and they are non-polar because they do not have a polar axis. Importantly, these classes can form crystals that are optically active if the molecular arrangement is chiral, but the crystal class itself is not chiral. This matches the requirement for optical activity in achiral, non-polar crystals.

Evaluating the options, **E** is the correct choice because it includes crystal classes with no symmetry elements that would make them chiral and are non-polar. The other options either include symmetry elements that could lead to chirality or are not consistently achiral and non-polar.

**Answer: E**
\end{tracebox}

\textit{Interpretation.} This example shows the main tradeoff in the external evaluation. The trained model is much more structured and decisive, but that additional answer discipline does not guarantee correctness.

\end{document}